\documentclass[10pt,journal,compsoc]{IEEEtran}

\usepackage{cite}
\usepackage{times}
\usepackage{epsfig}
\usepackage{graphicx}
\usepackage{amsmath}
\usepackage{amssymb}
\usepackage{multirow}
\usepackage{amsthm}
\usepackage{subfigure}
\usepackage{color}
\usepackage{csquotes}
\usepackage{soul}
\usepackage{enumerate}
\usepackage[normalem]{ulem}
\usepackage{algorithm}
\usepackage{algorithmicx}
\usepackage{algpseudocode}
\usepackage{booktabs}
\usepackage{array}
\usepackage{rotating}
\usepackage{arydshln}
\usepackage{ragged2e}
\usepackage{hyperref}
\hypersetup{pagebackref=false,breaklinks=true,colorlinks,bookmarks=false}

\newcommand\eg{\emph{e.g.}} 
\newcommand\ie{\emph{i.e.}}

\newcommand\etal{\emph{et al.}}

\begin{document}
%
\title{Uncertainty Inspired RGB-D Saliency Detection
}
%
%
%
%

       
\author{Jing~Zhang,
       Deng-Ping~Fan,
       Yuchao~Dai, \\
       Saeed~Anwar, 
       Fatemeh~Saleh, 
       Sadegh~Aliakbarian,
       and
       Nick~Barnes
\IEEEcompsocitemizethanks{\IEEEcompsocthanksitem Jing Zhang is with Research School of Engineering, the Australian National University, ACRV, DATA61-CSIRO. (Email: zjnwpu@gmail.com)
\IEEEcompsocthanksitem Deng-Ping Fan is with the CS, Nankai University, China. (Email: dengpfan@gmail.com)
\IEEEcompsocthanksitem Yuchao Dai is with School of Electronics and Information, Northwestern Polytechnical University, China. (Email: daiyuchao@gmail.com)
\IEEEcompsocthanksitem Saeed Anwar is with the Australian National University, DATA61-CSIRO. (Email: saeed.anwar@data61.csiro.au)
\IEEEcompsocthanksitem Fatemeh Saleh is with the Australian National University, ACRV. (Email: fatemehsadat.saleh@anu.edu.au)
\IEEEcompsocthanksitem Sadegh Aliakbarian is with the Australian National University, ACRV. (Email: sadegh.aliakbarian@anu.edu.au)
\IEEEcompsocthanksitem Nick Barnes is with Research School of Engineering, the Australian National University. (Email: nick.barnes@anu.edu.au)
\IEEEcompsocthanksitem A preliminary version of this work appeared at CVPR 2020 \cite{jing2020uc}. 
\IEEEcompsocthanksitem Corresponding author: Deng-Ping Fan.


}
}

%
%

\markboth{Journal of \LaTeX\ Class Files,~Vol.~14, No.~8, August~2015}%
{Shell \MakeLowercase{\textit{et al.}}: Bare Demo of IEEEtran.cls for Computer Society Journals}
%



\IEEEtitleabstractindextext{%
\begin{abstract}
\justifying
We propose the first stochastic framework to employ uncertainty for RGB-D saliency detection by learning from the data labeling process. Existing RGB-D saliency detection models treat this task as a point estimation problem by predicting a single saliency map following a deterministic learning pipeline. We argue that, however, the deterministic solution is relatively ill-posed.  
Inspired by the saliency data labeling process, we propose a generative architecture to achieve probabilistic RGB-D saliency detection which utilizes a latent variable to model the labeling variations. Our framework includes two main models:
1) a generator model, which maps the input image and latent variable to stochastic saliency prediction,
and 2) an inference model, which gradually updates the latent variable by sampling it from the true or approximate posterior distribution. The generator model is an encoder-decoder saliency network. To infer the latent variable, we introduce two different solutions: i) a Conditional Variational Auto-encoder with an extra encoder to approximate the posterior distribution of the latent variable; and ii)
an Alternating Back-Propagation technique, which directly samples the latent variable from the true posterior distribution.
Qualitative and quantitative results on six challenging RGB-D benchmark datasets show our approach's superior performance in learning the distribution of saliency maps. The source code is
publicly available via our project page: \url{https://github.com/JingZhang617/UCNet}.

\end{abstract}

}

\maketitle

\IEEEdisplaynontitleabstractindextext

\begin{figure*}[t!]
   \begin{center}
   {\includegraphics[width=0.99\linewidth]{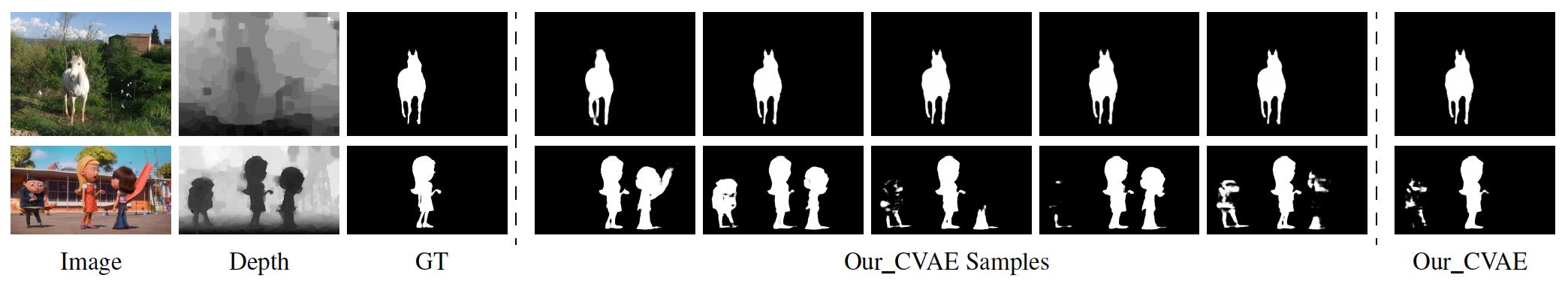}}  
   \end{center}
  \vspace{-2mm}
   \caption{GT compared with our predicted saliency maps.
   For simple context image (first row), we can produce consistent predictions. When dealing with complex scenarios where there exists uncertainties in salient regions (second row), our model can produce diverse predictions (\enquote{Our\_CVAE Samples}), where \enquote{Our\_CVAE} is our deterministic prediction after the saliency consensus module, which will be introduced in Section \ref{determnistic_inference_sec}.
   }
   \label{fig:inconsistent_ef_sod}
\end{figure*}

\IEEEpeerreviewmaketitle

\IEEEraisesectionheading{\section{Introduction}\label{sec:introduction}}

%
%
%
%

\IEEEPARstart{O} {bject}-level saliency detection (\ie, salient object detection) involves separating the most conspicuous objects that attract human attention from the background \cite{itti_saliency,achanta2009frequency,Iter_Coop_CVPR,Zhang_2018_CVPR,Liu_2019_ICCV,F3Net_aaai2020,jing2020weakly,SCRN_iccv}. Recently, visual saliency detection from RGB-D images has attracted lots of interests due to the importance of depth information in the
human vision system and the popularity of depth sensing technologies \cite{dmra_iccv19,zhao2019Contrast,select_focus_rgbd,self_attention_rgbd,JLDCF_rgbd,a2dele_rgbd,jing2020uc}. With the extra depth data,
conventional RGB-D saliency detection models focus on predicting one single saliency map
for
the RGB-D input by exploring the complementary information between the RGB image and the depth data.


The standard practice for RGB-D saliency detection is to train a deep neural network using ground-truth (GT) saliency maps provided by the corresponding benchmark datasets, 
thus formulating saliency detection as a point estimation problem by learning a mapping function $Y=f(X;\theta)$, where $\theta$ represents network parameter set, and $X$ and $Y$ are input RGB-D image pair and corresponding GT saliency map.
Usually, the GT saliency maps are obtained through human consensus or by the dataset creators \cite{sip_dataset}. 
Building upon large scale RGB-D datasets, deep convolutional neural network-based RGB-D saliency detection models \cite{JLDCF_rgbd,dmra_iccv19,chen2019three,han2017cnns,zhao2019Contrast} have made profound progress.
We argue that the way RGB-D saliency detection progresses
through the conventional pipelines \cite{JLDCF_rgbd,dmra_iccv19,chen2019three,han2017cnns,zhao2019Contrast} fails to capture the uncertainty in labeling the GT saliency maps.

According to research in human visual perception \cite{scanpath}, visual saliency detection is subjective to some extent. Each person could have specific preferences \cite{henderson2017meaning}
in labeling the saliency map (which has been discussed in user-specific saliency detection \cite{ITTI20001489}). More precisely speaking, the GT labeling process is never a deterministic process, which is
different from 
category-aware tasks, such as
semantic segmentation \cite{FCN}, as a \enquote{Table} will never be ambiguously labeled as \enquote{Cat}, while the salient foreground for one annotator may be defined as background by other annotators as shown in the second row of Fig.~\ref{fig:inconsistent_ef_sod}.

In Fig.~\ref{fig:inconsistent_ef_sod}, we present the
GT saliency map and other candidate salient regions (produced by our CVAE-based method, which will be introduced in Section \ref{rgbd_cvae}) that may attract human attention. Fig.~\ref{fig:inconsistent_ef_sod} shows that the deterministic mapping (from \enquote{Image} to \enquote{GT}) may lead to an \enquote{over-confident} model, as the provided \enquote{GT} may be biased
as shown in the second row of Fig.~\ref{fig:inconsistent_ef_sod}.
To overcome this, instead of performing point estimation, we are interested in how the network achieves distribution estimation with diverse saliency maps produced\footnote{Diversity of predictions depends on the context of the image, where simple context images will lead to consistent predictions, and complex context images may generate diverse predictions.}, capturing the uncertainty of human annotation.
Furthermore, in practice, it is more desirable to have multiple saliency maps produced to reflect human uncertainty instead of a single saliency map prediction for subsequent tasks.

Inspired by human perceptual uncertainty, as well as the labeling process of saliency maps, we propose a generative architecture to achieve
probabilistic RGB-D saliency detection with a latent variable $z$ modeling human uncertainty in the annotation. 
Two main models are included in this framework: 1) a generator (\ie, encoder-decoder) model, which maps the input RGB-D data and latent variable to stochastic saliency prediction; and 2) an inference model, which progressively refreshes the latent variable. 
To infer the latent variable, we introduce two different strategies: 
\begin{itemize}
\item 
A Conditional Variational Auto-encoder (CVAE)~\cite{structure_output} based model with an additional encoder to approximate the posterior distribution of the latent variable.

\item
The Alternating Back-Propagation (ABP)~\cite{ABP_aaai} based technique, which directly samples the latent variable from the true posterior distribution via Langevin Dynamics based Markov chain Monte Carlo (MCMC) sampling~\cite{langevin_mcmc,Neal2010MCMCUH}.
\end{itemize}


This paper is an extended version of our conference paper, UC-Net~\cite{jing2020uc}. In particular, UC-Net focuses on generating saliency maps via CVAE and augmented ground-truth to model diversity and to avoid posterior collapse problem~\cite{Lagging_Inference_Networks}. While UC-Net showed promising performance by modeling such variations, it still has a number of shortcomings. Firstly, UC-Net requires engineering efforts (ground-truth augmentation) to model diversity and achieve stabilized training (mitigating posterior collapse). Here, we use a simpler technique to achieve the same goal, by using the standard KL-annealing strategy~\cite{kl_annealing1,beta_vae} with less human intervention. Experimental results in Fig. \ref{fig:variance_maps} clearly illustrate the effectiveness of the KL-annealing strategy. Secondly, we improve the quality of the generated saliency maps by designing a more expressive decoder that benefits from spatial and channel attention mechanisms~\cite{fu2019dual}. Thirdly, inspired by~\cite{structure_output} we modify the cost function of UC-Net to reduce the discrepancy in encoding the latent variable at training and test time, which is elaborated in Section~\ref{sec:OurApproach}.

Moreover, CVAE-based methods approximate the posterior distribution via an inference model (or an encoder) and optimize the evidence lower bound (ELBO). The lower bound is simply the composition of the reconstruction loss and the divergence between the approximate posterior and prior distribution. If the model focuses more on optimizing the reconstruction quality, the latent space may fail to learn meaningful representation. On the other hand, if the model focuses more on reducing the divergence between the approximate posterior and prior distribution, the model may sacrifice the reconstruction quality. Additionally, since the model approximates the posterior distribution rather than modeling the true posterior, it may lose expressivity in general. Here, we propose to use Alternating Back-Propagation (ABP) technique~\cite{ABP_aaai} that directly samples latent variables from the true posterior. While it is much simpler, our experimental results show ABP leads to impressive result for generating saliency maps. Note that both CVAE-based and ABP-based solutions can produce stochastic saliency predictions by modeling output space distribution as a generative model conditioned on the input RGB-D image pair. Similar to UC-Net, during the testing phase, a saliency consensus module is introduced to mimic the majority voting mechanism for GT saliency map generation, and generate one single saliency map in the end for performance evaluation. Finally, in addition to producing state-of-the-art results, our experiments provide a thorough evaluation of the different components of our model as well as an extensive study on the diversity of the generated saliency maps.

Our main contributions are summarized as: 1) We propose the first uncertainty inspired probabilistic RGB-D saliency prediction model with a latent variable $z$ introduced to the network to represent human uncertainty in annotation;
2) We introduce two different schemes to infer the latent variable, including a CVAE \cite{structure_output} framework with an additional encoder to approximate the posterior distribution of $z$ and an ABP~\cite{ABP_aaai} pipeline, which samples the latent variable directly from its true posterior distribution via Langevin dynamics based Markov chain Monte Carlo (MCMC) sampling \cite{Neal2010MCMCUH}.
Each of them can model the conditional distribution of output, and lead to diverse predictions during testing;
3) Extensive experimental results on six RGB-D saliency detection benchmark datasets demonstrate the effectiveness of our proposed solutions.



\section{Related Work}
In this section, we first briefly review existing RGB-D saliency detection models. We then investigate existing generative models, including Variational Auto-encoder (VAE) \cite{vae_bayes_kumar,structure_output}, and Generative Adversarial Networks (GAN) \cite{GAN_nips,conditional_gan}. We also highlight the
uniqueness of the proposed solutions in this section.

\subsection{RGB-D Saliency Detection}
Depending on how the complementary information of RGB images and depth data is fused, existing RGB-D saliency detection models can be roughly classified into three categories: early-fusion models \cite{qu2017rgbd,jing2020uc}, late-fusion models \cite{wang2019adaptive,han2017cnns} and cross-level fusion models \cite{dmra_iccv19,chen2018progressively,chen2019multi,chen2019three,zhao2019Contrast,select_focus_rgbd,self_attention_rgbd,JLDCF_rgbd,fu2020siamese,a2dele_rgbd,fan2020bbs,zhai2020bifurcated,ji2020accurate,pang2020hierarchical,zhang2020bilateral}. 
The first solution directly concatenates the RGB image with its depth information, forming a four-channel input, and feed it to the network to obtain both the appearance information and geometric information.
\cite{qu2017rgbd} proposed an early-fusion model to generate features
for each superpixel of the RGB-D pair, which was then fed to a CNN to produce saliency of each superpixel. The second approach treats each modality independently, and predictions from both modalities are fused at the end of the network.
\cite{wang2019adaptive} introduced a late-fusion network (\ie, AFNet) to fuse predictions from the RGB and depth branch adaptively. In a similar pipeline,
\cite{han2017cnns} fused the RGB and depth information through fully connected layers.
The third one fuses intermediate features of each modality by considering correlations of the above two modalities.
To achieve this,
\cite{chen2018progressively} presented a complementary-aware fusion block.
\cite{chen2019three} designed attention-aware cross-level combination blocks to obtain complementary information of each modality.
\cite{zhao2019Contrast}
employed a fluid pyramid integration framework to achieve multi-scale cross-modal feature fusion.
\cite{self_attention_rgbd} designed a self-mutual attention model to effectively fuse RGB and depth information. Similarly,
\cite{select_focus_rgbd} presented a complimentary interaction module (CIM) to select complementary representation from the RGB and depth data.
\cite{JLDCF_rgbd} provided joint learning and densely-cooperative fusion framework for complementary feature discovery.
\cite{a2dele_rgbd} introduced a
depth distiller to transfer the depth knowledge from the depth stream to the RGB stream to achieve a lightweight architecture without use of depth data at test time. A comprehensive survey can be found in ~\cite{zhou2020rgb}.
\subsection{VAE or CVAE-based Deep Probabilistic Models}
Ever since the seminal work by Kingma \etal \cite{vae_bayes_kumar} and Rezende \etal \cite{pmlr-v32-rezende14}, VAE and its conditional counterpart CVAE \cite{structure_output} have been widely applied in various computer vision problems. 
A typical VAE-based model consists of an encoder, a decoder, and a loss function. The encoder is a neural network with weights and biases $\theta$, which maps the input datapoint $X$ to a latent (hidden) representation $z$. The decoder is another neural network with weights and biases $\phi$, which reconstructs the datapoint $X$ from $z$. 
To train a VAE, a reconstruction loss and a regularizer are needed to penalize the disagreement of the latent representation's prior and posterior distribution. 
Instead of defining the prior distribution of the latent representation as a standard Gaussian distribution, CVAE-based networks
utilize the input observation to modulate the prior on Gaussian latent variables to generate the output.

In low-level vision, VAE and CVAE have been applied to 
tasks such as 
latent representations with sharp samples~\cite{pixel_vae}, difference of motion modes~\cite{MT-VAE}, medical image segmentation models
\cite{PHiSeg2019}, and modeling inherent ambiguities of an image~\cite{probabilistic_unet}.
Meanwhile, VAE and CVAE have been explored in more complex vision tasks such as uncertain future forecast~\cite{vae_future}, salient feature enhancement~\cite{ContrastiveVAE}, human motion prediction~\cite{aliakbarian2019learning,aliakbarian2019sampling}, and shape-guided image generation~\cite{Esser_2018_CVPR}. Recently, VAE and CVAE have been extended to 3D domain targeting applications such as 3D meshes deformation~\cite{Tan_2018_CVPR}, and point cloud instance segmentation~\cite{Yi_2019_CVPR}. For saliency detection, \cite{SuperVAE_AAAI19} adopted VAE to model image background, and separated salient objects from the background through the reconstruction residuals.


\subsection{GAN or CGAN-based Dense Models}
GAN \cite{GAN_nips} and its conditional counterparts \cite{conditional_gan} have also been used in dense prediction tasks. Existing GAN-based dense prediction models mainly focus on two directions: 1) using GANs
in a
fully supervised manner \cite{semantic_gan_nips_2016,SegGAN,SegAN_2017, Sal_CGAN,cascaded_conv_adv_sod} and treat the discriminator loss as a higher-order regularizer for dense prediction; or 2) apply GANs
to 
`semi-supervised scenarios \cite{Hung_semiseg_2018,semi_seg_gan}, where the output of the discriminator serves as guidance to evaluate the degree of the unsupervised sample participating in network training. 
In saliency detection, following the first direction, \cite{Pan_2017_SalGAN} introduced a discriminator in the fixation prediction network to distinguish predicted fixation map and ground-truth. Different from the above two directions, \cite{Jiang2019emphcmSalGANRS} adopted GAN in a RGB-D saliency detection network to explore the intra-modality (RGB, depth) and cross-modality simultaneously. \cite{DSAL-GAN} used GAN as a denoising technique to clear up the noisy input images.
\cite{cascaded_conv_adv_sod} designed a discriminator to distinguish real saliency map (group truth) and fake saliency map (prediction), thus structural information can be learned without CRF \cite{dense_crf} as post-processing technique. \cite{multi_spectual_sod_adv} adopted CycleGAN \cite{cycle_gan} as an domain adaption technique to generate pseudo-NIR image for existing RGB saliency dataset and achieve multi-spectral image salient object detection.


\subsection{Uniqueness of Our Solutions}
\label{unique_solution}
To the best of our knowledge, generative models have not been exploited in saliency detection to model annotation uncertainty, except for our preliminary version \cite{jing2020uc}.
As a conditional latent variable model, two different solutions can be used to infer the latent variable. One is CVAE-based~\cite{structure_output} method (the one we used in the preliminary version \cite{jing2020uc}), which infers the latent variable using Variational Inference, and another one is MCMC based method, which we propose to use in this work. 
Specifically, we present a new latent variable inference solution with less parameter load based on the alternating back-propagation technique \cite{ABP_aaai}.


CVAE-based models infer the latent variable through finding the ELBO of the log-likelihood to avoid MCMC as it was too slow in the non-deep-learning era. In other words, CVAEs approximates Maximum Likelihood Estimation (MLE) by finding the ELBO with an extra encoder. The main issue of this strategy is \enquote{posterior collapse} \cite{Lagging_Inference_Networks}, where the latent variable is independent of network prediction, making it unable to represent the uncertainty of human annotation. We introduced the \enquote{New Label Generation} strategy in our preliminary version \cite{jing2020uc} as an effective way to avoid posterior collapse problem. In this extended version, we propose a much simpler strategy using the KL annealing strategy\cite{kl_annealing1,beta_vae}, which slowly introduces the KL loss term to the loss function with an extra weight.
The experimental results show that this simple strategy can avoid the posterior collapse problem
with the provided single GT saliency map.

Besides the KL annealing term, we introduce ABP \cite{ABP_aaai} as an alternative solution to prevent
posterior collapse in the network.
ABP introduces
gradient-based MCMC and updates the latent variable with gradient descent back-propagation to directly train the network targeting MLE. Compared with CVAE, ABP samples latent variables directly from its true posterior distribution, making it more accurate in inferring the latent variable. Furthermore, no assistant network (the additional encoder in CVAE) used in ABP, which leads to smaller network parameter load. 

We introduce ABP-based inference model as an extension to the CVAE-based pipeline \cite{jing2020uc}. Experimental results show that both solutions can effectively estimate the latent variable, leading to stochastic saliency predictions. Details of the two inference models are
introduced in Section \ref{rgbd_cvae}.


\begin{figure}[t!]
   \begin{center}
   \begin{tabular}{ c@{ }}
   {\includegraphics[width=0.99\linewidth]{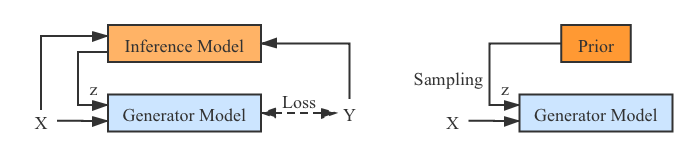}}\\
   \footnotesize{(a) Training pipeline}  \hspace{6em}  \footnotesize{(b) Testing pipeline} \\
   \end{tabular}
   \end{center}
   \caption{Training and testing pipeline. During training, the inferred latent variable $z$ and input image $X$ are fed to the \enquote{Generator Model} for stochastic saliency prediction. During testing, we sample from the prior distribution of $z$ to produce diverse predictions for each input image.
   }
   \label{fig:simplyfied_training_testing}
\end{figure}

\section{Our Model}\label{sec:OurApproach}
In this section, we present our probabilistic RGB-D saliency detection model, which learns the underlying conditional distribution of saliency maps rather than a mapping function from RGB-D input to a single saliency map.
Let $\mathcal{D} = \{X_i,Y_i\}_{i=1}^N$ be the training dataset, where $X_i$ denotes the RGB-D input,
$Y_i$ denotes the GT saliency map, and $N$ denotes the total number of images in the dataset.
We intend to model $P_\omega(Y|X,z)$, where $z$ is a latent variable representing the inherent uncertainty in salient regions which can be also seen in how a human annotates salient objects. Our framework utilizes two main components during training:
1) a generator model, which maps input RGB-D $X$ and latent variable $z$ to conditional prediction $P_\omega(Y|X,z)$; and 2) an inference model, which infers the latent variable $z$. During testing, we can sample multiple latent variables from the learned prior distribution $P_\theta(z|X)$
to produce stochastic saliency prediction. The whole pipeline of our model during training and testing is illustrated in Fig.~\ref{fig:simplyfied_training_testing} (a) and (b) respectively. Specifically, during training, the model learns saliency from the \enquote{Generator Model}, and updates the latent variable with the \enquote{Inference Model}. During testing, we sample from the \enquote{Prior} distribution of the latent variable to obtain stochastic saliency predictions.

\begin{figure}[tbp]
   \includegraphics[height=0.65\linewidth]{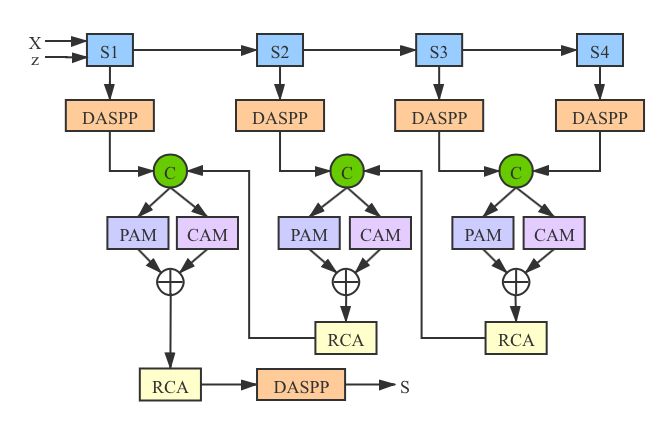}
   \caption{Details of the \enquote{Generator Model}, which takes image $X$ and latent variable $z$ as input, and produce stochastic saliency map $S$, where \enquote{S1-S4} represent the four convolutional blocks of our backbone network. \enquote{DASPP} is the DenseASPP module \cite{denseaspp}, \enquote{PAM} and \enquote{CAM} are position attention and channel attention module \cite{fu2019dual}, \enquote{RCA} is the Residual Channel Attention operation from \cite{zhang2018rcan}.
   }
   \label{fig:saliency_feature_net}
\end{figure}

\subsection{Generator Model}
The Generator Model takes $X$ and latent variable $z$ as input, and produces stochastic prediction $S=P_\omega(Y|X,z)$, where $\omega$ is the parameter set of the generator model. We choose ResNet50 \cite{ResHe2015} as our backbone, which contains four convolutional blocks.
To enlarge the receptive field, we follow DenseASPP~\cite{denseaspp} to obtain a
feature map with the receptive field of the whole image on each stage of the backbone network. We then gradually concatenate the two adjacent feature maps in a top-down manner and feed it to a \enquote{Residual Channel Attention} module \cite{zhang2018rcan} to obtain stochastic saliency map $S$.
As illustrated in Fig. \ref{fig:saliency_feature_net}, our generator model follows the recent progress
in dense prediction problems such as semantic segmentation~\cite{FCN}, via a proper use of a hybrid attention mechanism. To this end, our generator model benefits from two types of attention: a Position Attention Module~\cite{fu2019dual} and a Channel Attention Module~\cite{fu2019dual}. The former aims to capture the spatial dependencies between any two locations of the feature map, while the latter aims to capture the channel dependencies between any two channel in the feature map. We follow~\cite{fu2019dual} to aggregate and fuse the outputs of these two attention modules to further enhance the feature representations.

\subsection{Inference Model}
\label{rgbd_cvae}
We propose two different solutions to infer or update the latent variable $z$: 1) A CVAE-based~\cite{structure_output} pipeline, in which we approximate the posterior distribution via a neural network (\ie, the encoder);
and 2) An ABP~\cite{ABP_aaai} based strategy to sample directly from the true posterior distribution of $z$ via Langevin Dynamics based MCMC \cite{langevin_mcmc}.

\noindent\textbf{Infer $z$ with CVAE:}
The Variational Auto-encoder \cite{vae_bayes_kumar} is
a directed graphical model and typically comprise of two fundamental components, an encoder that maps the input variable $X$ to the latent space $Q_\phi(z|X)$, where $z$ is a low dimensional Gaussian variable and a decoder that reconstructs $X$ from $z$ to get $P_\omega(X|z)$. To train the VAE, a reconstruction loss and a regularizer 
to 
penalize the disagreement of the prior and the approximate posterior distribution of $z$ are
utilized as:
\begin{equation}
\label{vae_equation}
\begin{aligned}
    \mathcal{L}_{\mathrm{VAE}} = E_{z\sim Q_\phi(z|X)}[-\log P_\omega(X|z)] \\
    +D_{KL}(Q_\phi(z|X)||P(z)),
\end{aligned}
\end{equation}
where the first term is the reconstruction loss, or the expected negative log-likelihood, and the second term is a regularizer, which is Kullback-Leibler divergence $D_{KL}(Q_\phi(z|X)||P(z))$ to reduce the gap between the normally distributed prior $P(z)$ and the approximate posterior $Q_\phi(z|X)$. The expectation $E_{z\sim Q_\phi(z|X)}$ is taken with the latent variable $z$ generated from the approximate posterior distribution $Q_\phi(z|X)$.




Different from the VAE, which model marginal likelihood ($P(X)$ in particular) with a
latent variable generated from the standard normal distribution, the CVAE \cite{structure_output} modulates the prior of latent variable $z$ as a Gaussian distribution with parameters conditioned on the input data $X$. There are three types of variables in the conditional generative model:
conditioning variable, latent variable, and output variable. In our saliency detection scenario, we define output as the saliency prediction $Y$, and latent variable as $z$. As our output $Y$ is conditioned
on the input RGB-D data $X$, we then define the input $X$ as the
conditioning variable.  
For the latent variable $z$ drawn from the Gaussian distribution $P_\theta(z|X)$, the output variable $Y$ is generated from $P_\omega(Y|X,z)$,
then the posterior of $z$ is formulated as $Q_\phi(z|X,Y)$, representing feature embedding of the given input-output pair $(X,Y)$.

\begin{figure}[t!]
   \begin{center}
   {\includegraphics[width=0.95\linewidth]{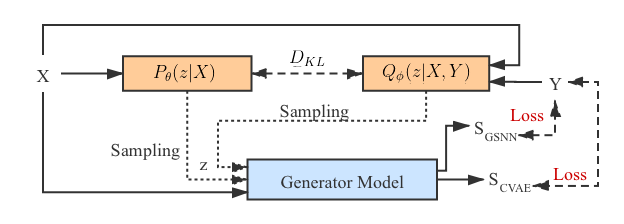}}
   \end{center}
   \caption{RGB-D saliency detection via CAVE. The \enquote{Generator Model} is shown in Fig. \ref{fig:saliency_feature_net}. During training, we sample from both posterior net $z\sim Q_\phi(z|X,Y)$ and prior net $z\sim P_\theta(z|X)$ to obtain predictions $S_{CVAE}$ and $S_{GSNN}$ respectively. During testing, $S_{GSNN}$ is our prediction.
   }
   \label{fig:cvae_overview} 
\end{figure}

The loss of CVAE is defined as:
\begin{equation}
\label{CVAE_equation}
\begin{aligned}
    \mathcal{L}_{\mathrm{CVAE}} = E_{z\sim Q_\phi(z|X,Y)}[-\log P_\omega(Y|X,z)] \\
    + \lambda_{kl}*D_{KL}(Q_\phi(z|X,Y)||P_\theta(z|X)),
\end{aligned}
\end{equation}
where $P_\omega(Y|X,z)$ is the likelihood of $P(Y)$ given latent variable $z$ and conditioning variable $X$, the Kullback-Leibler divergence $D_{KL}(Q_\phi(z|X,Y)||P_\theta(z|X))$ works as a regularization loss to reduce the gap between the prior $P_\theta(z|X)$ and the auxiliary posterior $Q_\phi(z|X,Y)$. Furthermore, to prevent the possible \textit{posterior collapse} problem as mentioned in Section \ref{unique_solution}, we introduce a
linear KL annealing \cite{kl_annealing1,beta_vae} term $\lambda_{kl}$ as weight for the KL loss term $D_{KL}$, which is defined as $\lambda_{kl}=ep/N_{ep}$,
where $ep$ is current epoch, and $N_{ep}$ is the maximum epoch number.
In this way, during training, the CVAE aims to model the conditional log likelihood of prediction under encoding error $D_{KL}(Q_\phi(z|X,Y)||P_\theta(z|X))$. During testing, we can sample from the prior network $P_\theta(z|X)$ to obtain stochastic predictions. 

As explained in \cite{structure_output}, the conditional auto-encoding of output variables at training may not be optimal to make predictions at test time, as the CVAE
uses a
posterior of $z$ ($z\sim Q_\phi(z|X,Y)$) for the reconstruction loss in the training stage, while it
uses the prior of $z$ ($z\sim P_\theta(z|X)$) during testing. One solution to mitigate the discrepancy in encoding the latent variable at training and testing is to allocate more weights to the KL loss term (\eg, $\lambda_{kl}$).
Another solution is setting the posterior network the
same as the prior network, \ie, $Q_\phi(z|X,Y)=P_\theta(z|X)$, and we can sample the latent variable $z$ directly from prior network in both training and testing stages.
We call this model the \enquote{Gaussian Stochastic Neural Network} (GSNN) \cite{structure_output}, and the objective function is: 
\begin{equation}
    \label{gsnn_equ}
    \begin{aligned}
    \mathcal{L}_{\mathrm{GSNN}} = E_{z\sim P_\theta(z|X)}[-\log P_\omega(Y|X,z)].
\end{aligned}
\end{equation}

We can combine the two objective functions introduced above ($\mathcal{L}_{\mathrm{CVAE}}$ and $\mathcal{L}_{\mathrm{GSNN}}$) to obtain a hybrid objective function:

\begin{equation}
    \label{hybrid_func}
    \mathcal{L}_{\mathrm{Hybrid}} = \alpha \mathcal{L}_{\mathrm{CVAE}}+(1-\alpha)\mathcal{L}_{\mathrm{GSNN}}
\end{equation}

\begin{figure}[t!]
   \begin{center}
   {\includegraphics[width=0.98\linewidth]{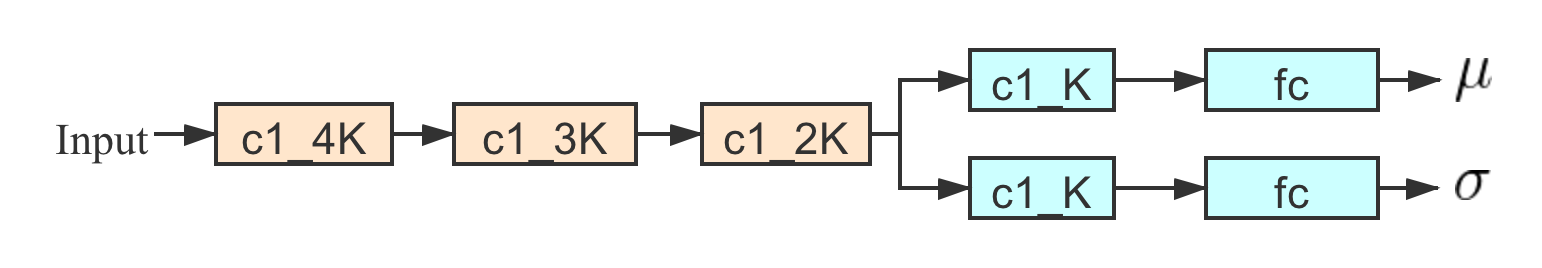}}
   \end{center}
   \caption{Detailed structure of inference models, where $K$ is dimension of the latent space, \enquote{c1\_4K} represents a $1 \times 1$ convolutional layer of output channel size $4\times K$, \enquote{fc} represents the fully connected layer.}
   \label{fig:encoder_latent}
\end{figure}

Following the standard practice of CVAE \cite{structure_output}, we design a CVAE-based RGB-D saliency detection pipeline as shown in Fig. \ref{fig:cvae_overview}. The two inference models ($Q_\phi(z|X,Y)$ and $P_\theta(z|X)$) share same structure as shown in Fig. \ref{fig:encoder_latent}, except for $Q_\phi(z|X,Y)$, we have concatenation of $X$ and $Y$ as input, while $P_\theta(z|X)$ takes $X$ as input.
Let's define $P_\theta(z|X)$ as PriorNet, which maps the input RGB-D data $X$ to a low-dimensional latent feature space, where $\theta$ is the parameter set of PriorNet. With the provided GT saliency map $Y$, we define $Q_\phi(z|X, Y)$ as PosteriorNet, with $\phi$ being the network parameter set.
We use five convolutional layers and two fully connected layers to map the input RGB-D image $X$ (or concatenation of $X$ and $Y$ for
PosteriorNet) to the statistics of the latent space: $(\mu_{\mathrm{prior}}, \sigma_{\mathrm{prior}})$ for 
PriorNet and $(\mu_{\mathrm{post}}, \sigma_{\mathrm{post}})$ for 
PosteriorNet respectively. Then the corresponding latent vector $z$ can be achieved with the reparameteration trick: $z=\mu+\sigma \cdot \epsilon$, where $\epsilon\sim\mathcal{N}(0,\mathbf{I})$.



According to Eq. \ref{hybrid_func}, the KL-divergence in $\mathcal{L}_{\mathrm{CVAE}}$ is used to measure the distribution mismatch between the $P_\theta(z|X)$ and $Q_\phi(z|X,Y)$, or how much information is lost when using $Q_\phi(z|X,Y)$ to represent $P_\theta(z|X)$. The GSNN loss term $\mathcal{L}_{\mathrm{GSNN}}$, on the other hand, can mitigate the discrepancy in encoding the latent variable during training and testing. The hybrid loss in Eq. \ref{hybrid_func} can achieve structured outputs with hyper-parameter $\alpha$ to balance the two objective functions in Eq.~\ref{CVAE_equation} and Eq.~\ref{gsnn_equ}. 
\noindent\textbf{Infer $z$ with ABP:} As mentioned earlier, one drawback of CVAE-based models is the posterior collapse problem \cite{Lagging_Inference_Networks}, where the model learns to ignore the latent variable, thus it becomes independent of the prediction $Y$, as $Q_\phi(z|X,Y)$ will simply collapse to $P_\theta(z|X)$, and $z$ embeds no information about the prediction. In our scenario, the \enquote{Posterior Collapse} phenomenon can be interpreted as the fact that the latent variable $z$ fails to capture the inherent human uncertainty in the
annotations. To this end, we propose another alternative solution based on alternating back-propagation \cite{ABP_aaai}. Instead of approximating
the posterior of $z$ with an
encoder network as in a CVAE, we directly sample $z$ from its true posterior distribution via gradient based MCMC.

Alternating Back-Propagation \cite{ABP_aaai} was introduced for learning the generator network model. It updates the latent variable and network parameters in an EM-manner. Firstly, given network prediction with the
current parameter set, it infers the latent variable by
Langevin dynamics based MCMC, which they call 
\enquote{Inferential back-propagation} \cite{ABP_aaai}. Secondly, given the updated latent variable, the network parameter set is updated with gradient descent, and they call it 
\enquote{Learning back-propagation} \cite{ABP_aaai}.
Following the previous variable definitions,
given the training example $(X,Y)$, we intend to infer $z$ and learn the network parameter $\omega$ to minimize the reconstruction error as well as a regularization term that corresponds to the prior on $z$.

\begin{algorithm}[!t]
\small
\caption{Learning Stochastic Saliency via Alternating Back-propagation}
\label{alg:abp_algorithm}
\textbf{Input}: Training dataset $D = \{(X_i,Y_i)\}_{i=1}^N$

\textbf{Network Setup}: Maximal epoch $N_{ep}$, number of Langvin steps $l$, step size $s$, learning rate $\gamma$

\textbf{Output}: Network parameter set $\omega$ and the inferred latent variable $\{z_i\}_{i=1}^N$

\begin{algorithmic}[1]
\State Initialize backbone of the \enquote{Generator Model} with ResNet50 \cite{ResHe2015} for image classification, and other new added layers with a truncated Gaussian distribution. Initialize $z_i$ with standard Gaussian distribution.
\For{$t = 1,...,N_{ep}$}
\State \textbf{Inferential back-propagation}: For each $i$, run $l$ steps of Langevin Dynamics to sample $z_i \sim P_{\omega}(z_i|Y_i,X_i)$ following Eq. \ref{langevin_dynamics}, with $z_i$ initialized as Gaussian white noise (first iteration) or obtained from previous iteration.
\State \textbf{Learning back-propagation}: Update model parameters via: $\omega \leftarrow \omega + \gamma \frac{\partial \mathcal{L}(\omega)}{\partial \omega}$, where the gradient of $\mathcal{L}(\omega)$ can be obtained through stochastic gradient descent.
\EndFor
\end{algorithmic}
\end{algorithm}

As a non-linear generalization of factor analysis, the conditional generative model aims to generalize the mapping from continuous latent variable $z$ to the prediction $Y$ conditioned on the input image $X$. As in traditional factor analysis, we define our generative model as:
\begin{eqnarray}
    && z \sim P(z)=\mathcal{N}(0,\mathbf{I}), \label{eq:abp_1}\\
    && Y = f_\omega(X,z) + \epsilon, \epsilon \sim \mathcal{N}(0,\text{diag}(\sigma)^2), 
    \label{eq:abp_3} 
\end{eqnarray}
where $P(z)$ is the prior distribution of $z$. The conditional distribution of $Y$ given $X$ is $P_\omega(Y|X) = \int p(z) P_\omega(Y|X,z) dz$ with the latent variable $z$ integrated out. We define the observed-data log-likelihood as $L(\omega)=\sum_{i=1}^n \log P_\omega(Y_i|X_i)$, where the
gradient of $P_\omega(Y|X)$ is defined as:
\begin{equation}
\label{update_omega}
\begin{aligned}
    \frac{\partial}{\partial \omega}\log P_{\omega}(Y|X)&= 
    \frac{1}{P_{\omega}(Y|X)}\frac{\partial}{\partial \omega}  P_{\omega}(Y|X)\\
    &=\text{E}_{P_{\omega}(z|X,Y)} \left[\frac{\partial}{\partial \omega}\log P_{\omega}(Y,z|X)\right].
\end{aligned}   
\end{equation}

The expectation term $\text{E}_{P_{\omega}(z|X,Y)}$ can be approximated by drawing samples from $P_{\omega}(z|X,Y)$, and then computing
the Monte Carlo average. This step corresponds to inferring
the latent variable $z$. Following ABP \cite{ABP_aaai}, we use Langevin Dynamics based MCMC (a gradient-based Monte Carlo method) to sample $z$, which iterates:
\begin{equation}
\begin{aligned}
    z_{t+1}=z_{t}+ \frac{s^2}{2}\left[ \frac{\partial}{\partial z}\log P_{\omega}(Y,z_{t}|X)\right]+s \mathcal{N}(0,I_d),
    \label{langevin_dynamics}
\end{aligned}
\end{equation}
with 
\begin{equation}
\frac{\partial}{\partial z}\log P_{\omega}(Y,z|X) = \frac{1}{\sigma^2}(Y-f_\omega(X,z))\frac{\partial}{\partial z}f_\omega(X,z) - z,
\end{equation}
where $t$ is the
time step for Langevin sampling, and
$s$ is the step size. The whole pipeline of inferring latent variable $z$ via ABP is shown in Algorithm \ref{alg:abp_algorithm}.

\noindent\textbf{Analysis of two inference models:} Both the
CVAE-based~\cite{structure_output} inference model and ABP-based \cite{ABP_aaai} strategy can infer latent variable $z$, where the former one approximates the posterior distribution of $z$ with an extra encoder, while the latter solution targets at MLE by directly sampling from the true posterior distribution. As mentioned above, the CVAE-based solution may suffer from posterior collapse \cite{Lagging_Inference_Networks}, where the latent variable $z$ is independent of the prediction, making it unable to represent the uncertainty of labeling. To prevent 
posterior collapse, we adopt the KL annealing strategy \cite{kl_annealing1,beta_vae}, and let the KL loss term in Eq. \ref{CVAE_equation} gradually contribute to the CVAE loss function. On the contrary, the ABP-based solution suffers no posterior collapse problem,
which leads to simpler and more stable training,
where the latent variable $z$ is updated based on the current prediction. In both of our proposed solutions, with the inferred Gaussian random variable $z$, our model can
lead to stochastic prediction, with $z$ representing labeling variants.

\begin{figure}[!t]
   \begin{center}
   {\includegraphics[width=0.90\linewidth]{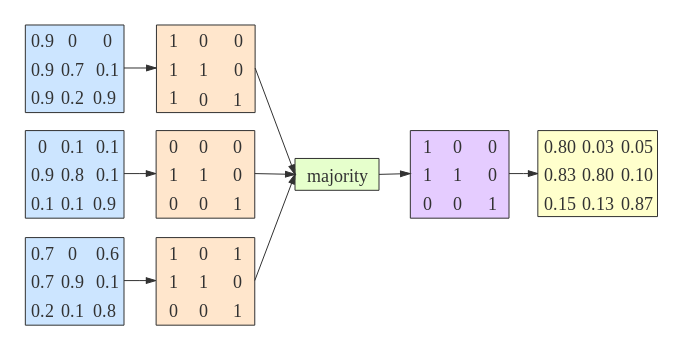}} 
   \end{center}
   \caption{Example showing how the saliency consensus module works.}
   \label{fig:majority_voting_example}
\end{figure}

\subsection{Output Estimation}
\label{determnistic_inference_sec}
Once the generative model parameters are learned, our model can produce prediction from
input $X$ following the generative process of the conditional generative model. With multiple iterations of sampling, we can obtain multiple saliency maps from
the same input $X$. To evaluate performance of the generative network, we need to estimate the deterministic prediction of the structured outputs. Inspired by \cite{structure_output}, our first solution is to simply average the multiple predictions. Alternatively, we can obtain multiple $z$ from the prior distribution, and define the deterministic prediction as $Y=f_\omega(X,E(z))$, where $E(z)$ is the mean of the multiple latent variable. Inspired by how the GT saliency map is obtained (\eg, Majority Voting), we introduce a
third solution, namely \enquote{Saliency Consensus Module}, which is introduced in detail.

\noindent\textbf{Saliency Consensus Module:}
To prepare a training dataset for saliency detection, multiple annotators are asked to label one image, and the majority \cite{sip_dataset} of saliency regions is defined as being salient in the final GT saliency map.

Although
the way in which the GT is acquired
is well known in the saliency detection community
yet, there exists no research on embedding this mechanism into deep saliency frameworks. The main reason is that current models define saliency detection as a point estimation problem instead of a distribution estimation problem, and the final single saliency map can not be further processed to achieve \enquote{majority voting}. We, instead, design a stochastic learning pipeline to obtain the conditional distributions of prediction, which makes it possible to perform a similar strategy as preparing the training data to generate deterministic prediction for performance evaluation. Thus, we introduce the saliency consensus module to compute the majority of different predictions in the testing stage as shown in Fig. \ref{fig:simplyfied_training_testing} (b).




During testing, we sample $z$ from PriorNet (for the CVAE-based inference model) or directly sample it from a
standard Gaussian distribution $\mathcal{N}(0,\mathbf{I})$, and feed it to the \enquote{Generator Model} to produce stochastic saliency prediction as shown in Fig. \ref{fig:simplyfied_training_testing} (b). With $C$ different samplings,
we can obtain $C$ predictions $P^1,..., P^C$.
We simultaneously feed these multiple predictions to the saliency consensus module to obtain the consensus of predictions for performance evaluation.

Given multiple predictions $\{P^c\}_{c=1}^C$, where $P^c \in [0,1]$, we first compute the binary\footnote{As the GT map $Y\in \{0,1\}$, we produce a
series of binary predictions with each one representing annotation from one saliency annotator.} version $P_b^c$ of the predictions by performing adaptive thresholding \cite{borji2015salient} on $P^c$.
For each pixel $(u,v)$, we obtain a $C$ dimensional feature vector $P_{u,v}\in\{0,1\}$. We define $P^{mjv}_b \in \{0,1\}$ as a one-channel saliency map representing the
majority of $P_{u,v}$, which is defined as:
\begin{equation}
    \label{majority_pred}
    \small
    P^{mjv}_b(u,v) = \left\{
    \begin{aligned}
    1 &, & \sum_{c=1}^C P_b^c(u,v)/C \geq 0.5,\\
    0 &, & \sum_{c=1}^C P_b^c(u,v)/C < 0.5.
    \end{aligned}
    \right.
\end{equation}
We define an indicator 
$\mathbf{1}^c(u,v)=\mathbf{1} (P^c_b(u,v)=P^{mjv}_b(u,v))$ representing whether the binary prediction is consistent with the majority of the predictions. If $P^c_b(u,v)=P^{mjv}_b(u,v)$, then $\mathbf{1}^c(u,v)=1$. Otherwise, $\mathbf{1}^c(u,v)=0$. We obtain one gray saliency map after saliency consensus as:
\begin{equation}
\label{majority_voting_equation}
{\small
    P^{mjv}_g(u,v)=\frac{\sum_{c=1}^C(P^c_b(u,v)\times \mathbf{1}^c(u,v))}{\sum_{c=1}^C \mathbf{1}^c(u,v)}.
}
\end{equation}


We show one toy example with $C=3$
in Fig. \ref{fig:majority_voting_example} to illustrate how the saliency consensus module works. As shown in Fig. \ref{fig:majority_voting_example}, given three gray-scale
predictions (illustrated in blue), we perform adaptive thresholding to obtain three different binary predictions (illustrated in orange). Then we compute a majority matrix (illustrated in purple), which is also binary, with each pixel representing majority prediction of the specific coordinate. Finally, after the saliency consensus module, our final gray-scale
prediction is computed based on mean of those pixels agreed (when $P^c_b(u,v)=P^{mjv}_b(u,v)$, we mean in location $u,v$, the prediction agrees with the majority) with the majority matrix,
and ignore others. For example, the majority of saliency in coordinate $(1,1)$ is 1, we obtain the gray prediction after the saliency consensus module as $(0.9+0.7)/2=0.8$, where 0.9 and 0.7 are predictions in $(1,1)$ of the first and third predictions.

\subsection{Loss function} 
\label{loss_func}
We introduce two different inference models to update the latent variable $z$: a CVAE-based model as shown in Fig. \ref{fig:cvae_overview}, and an ABP-based strategy as shown in Algorithm \ref{alg:abp_algorithm}. To further highlight structure accuracy of the prediction, we introduce smoothness loss based on the assumption that pixels inside a salient object should have a similar saliency value, and sharp distinction happens along object edges.

As an edge-aware loss, smoothness loss was initially introduced in \cite{UnsupeGodard} to encourage disparities to be locally smooth with an L1 penalty on the disparity gradients. It was then adopted in \cite{occlusion_aware} to recover optical flow in the occluded area by using an image prior.
We adopt smoothness loss to achieve a saliency map of high intra-class similarity, with consistent saliency prediction inside salient objects, and distinction happens along object edges.
Following ~\cite{occlusion_aware}, we define first-order derivatives of the saliency map in the smoothness term as
\begin{equation}
\label{smoothness_loss}
    \mathcal{L}_{\mathrm{Smooth}} = \sum_{u,v} \sum_{d\in{\overrightarrow{x},\overrightarrow{y}}} \Psi(|\partial_d P_{u,v}|e^{-\alpha |\partial_d Ig(u,v)|}),
\end{equation}
where $\Psi$ is defined as $\Psi(s) = \sqrt{s^2+1e^{-6}}$,
$P_{u,v}$ is the predicted saliency map at position $(u,v)$, and $Ig(u,v)$ is the image intensity, $d$ indexes over partial derivative in $\overrightarrow{x}$ and $\overrightarrow{y}$ directions. We set $\alpha=10$ in our experiments following the setting in \cite{occlusion_aware}.

We need to compute intensity $Ig$ of the image in the smoothness loss, as shown in Eq. \eqref{smoothness_loss}. To achieve this, we follow a saliency-preserving \cite{Saliency_preserving_eccv} color image transformation strategy and convert the RGB image $I$ to a gray-scale intensity image $Ig$ as:
\begin{equation}
    Ig = 0.2126\times I^{lr} + 0.7152\times I^{lg} + 0.0722\times I^{lb},
\end{equation}
where $I^{lr}$, $I^{lg}$, and $I^{lb}$ represent the color components in the linear color space after Gamma function be removed from the original color space. $I^{lr}$ is achieved via:
\begin{equation}
\label{gamma_extension}
\small
    I^{lr}=\left\{
\begin{aligned}
\frac{I^r}{12.92} & , & I^r\leq 0.04045, \\
\bigg(\frac{I^r+0.055}{1.055}\bigg)^{2.4} & , & I^r> 0.04045,
\end{aligned}
\right.
\end{equation}
where $I^r$ is the original red channel of image $I$, and we compute $I^g$ and $I^b$
in the same way as Eq. \eqref{gamma_extension}.

\noindent\textbf{CVAE Inference Model based Loss Function:}
For the CVAE-based inference model, we show its loss function in Eq. \ref{hybrid_func}, where the negative log-likelihood loss measures the reconstruction error. To preserve structure information and penalize wrong predictions along object boundaries, we adopt the structure-aware loss in \cite{F3Net_aaai2020}. The structure-aware loss is a weighted extension of cross-entropy loss, which integrates the boundary IOU loss \cite{Luo_2017_CVPR} to highlight the accuracy of boundary prediction.

With smoothness loss $\mathcal{L}_{\mathrm{Smooth}}$ and CVAE loss $\mathcal{L}_{\mathrm{Hybrid}}$, our final loss function for the CVAE-based framework is defined as:
\begin{equation}
\label{final_loss}
    \mathcal{L}_{\mathrm{sal}}^{CVAE} = \mathcal{L}_{\mathrm{Hybrid}}+\lambda_1\mathcal{L}_{\mathrm{Smooth}}.
\end{equation}
We tested $\lambda_1$ in the range of $[0.1, 0.2, \dots, 0.9, 1.0]$, and found ralatively better performance with $\lambda_1=0.3$.

\noindent\textbf{ABP Inference Model based Loss Function:} As there exists no extra encoder for the posterior distribution estimation, the loss function for the ABP inference model is simply the negative observed-data log-likelihood:
\begin{equation}
    \label{abp_loss_rec}
    \mathcal{L}_{ABP} = -\sum_{i=1}^n \log P_\omega(Y_i|X_i),
\end{equation}
which can be the same structure-aware loss as
in \cite{F3Net_aaai2020} similar to CVAE-based inference model.

Integrated with the above smoothness loss, we obtain the loss function for the ABP-based saliency detection model as:
\begin{equation}
    \label{abp_loss}
    \mathcal{L}_{\mathrm{sal}}^{ABP} = \mathcal{L}_{ABP} + \lambda_2\mathcal{L}_{\mathrm{Smooth}}.
\end{equation}
Similarly, we also empirically set $\lambda_2=0.3$ in our experiment.

\section{Experimental Results}
\label{experimental_res_sec}

\begin{table*}[t!]
  \centering
  \scriptsize
  \renewcommand{\arraystretch}{1.1}
  \renewcommand{\tabcolsep}{0.6mm}
  \caption{Benchmarking results of ten leading handcrafted feature-based models and eight deep models on six RGBD saliency datasets.  $\uparrow \& \downarrow$ denote larger and smaller is better, respectively. Here, we adopt mean $F_{\beta}$ and mean $E_{\xi}$\cite{Fan2018Enhanced}. Evaluation tool: \href{https://github.com/DengPingFan/D3NetBenchmark}{https://github.com/DengPingFan/D3NetBenchmark}.}
  \label{tab:BenchmarkResults}
  \begin{tabular}{lr|cccccccccc|cccccccc|ccc}
  \hline
  &  &\multicolumn{10}{c|}{Handcrafted Feature based Models}&\multicolumn{8}{c|}{Deep Models}&\multicolumn{3}{c}{Ours} \\
    & Metric &
   LHM  & CDB  & DESM & GP    &
   CDCP & ACSD & LBE & DCMC & MDSF   & SE   & DF   & AFNet& CTMF & MMCI & PCF   & TANet& CPFP & DMRA & UC-Net & CVAE & ABP \\
   &  & \cite{peng2014rgbd}        & \cite{liang2018stereoscopic}       & \cite{cheng2014depth}          & \cite{ren2015exploiting}              &
        \cite{zhu2017innovative}   & \cite{NJU2000}                 & \cite{feng2016local}  & \cite{cong2016saliency}
        & \cite{song2017depth}   & \cite{guo2016salient} &\cite{qu2017rgbd}       & \cite{wang2019adaptive}& \cite{han2017cnns}    & \cite{chen2019multi}
         & \cite{chen2018progressively}  &\cite{chen2019three}   &   \cite{zhao2019Contrast} & \cite{dmra_iccv19} &\cite{jing2020uc} & &  \\
  \hline
  \multirow{4}{*}{\textit{NJU2K}\cite{NJU2000}}
    & $S_{\alpha}\uparrow$    & .514 & .632 & .665 & .527 & .669 & .699 & .695 & .686 & .748 & .664 & .763 & .822 & .849 & .858 & .877 & .879 & .878 & .886 & .897 & \textbf{.902}& .900\\
    & $F_{\beta}\uparrow$     & .328 & .498 & .550 & .357 & .595 & .512 & .606 & .556 & .628 & .583 & .653 & .827 & .779 & .793 & .840 & .841 & .850 & .873 & .886 & \textbf{.893}& .889\\
    & $E_{\xi}\uparrow$       & .447 & .572 & .590 & .466 & .706 & .594 & .655 & .619 & .677 & .624 & .700 & .867 & .846 & .851 & .895 & .895 & .910 & .920 & .930 & \textbf{.937}& \textbf{.937}\\
    & $\mathcal{M}\downarrow$ & .205 & .199 & .283 & .211 & .180 & .202 & .153 & .172 & .157 & .169 & .140 & .077 & .085 & .079 & .059 & .061 & .053 & .051 & .043 & \textbf{.039} & \textbf{.039}\\ \hline
  \multirow{4}{*}{\textit{SSB}\cite{niu2012leveraging}}
    & $S_{\alpha}\uparrow$    & .562 & .615 & .642 & .588 & .713 & .692 & .660 & .731 & .728 & .708 & .757 & .825 & .848 & .873 & .875 & .871 & .879 & .835 & .903 & .898& \textbf{.904}\\
    & $F_{\beta}\uparrow$     & .378 & .489 & .519 & .405 & .638 & .478 & .501 & .590 & .527 & .611 & .617 & .806 & .758 & .813 & .818 & .828 & .841 & .837 & .884 & .878& \textbf{.886}\\
    & $E_{\xi}\uparrow$       & .484 & .561 & .579 &.508  & .751 & .592 & .601 & .655 & .614 & .664 & .692 & .872 & .841 & .873 & .887 & .893 & .911 & .879 & .938 & .935& \textbf{.939}\\
    & $\mathcal{M}\downarrow$ & .172 & .166 & .295 & .182 & .149 & .200 & .250 & .148 & .176 & .143 & .141 & .075 & .086 & .068 & .064 & .060 & .051 &.066 & .039 & .039& \textbf{.037}\\ \hline
  \multirow{4}{*}{\textit{DES}\cite{cheng2014depth}}
    & $S_{\alpha}\uparrow$    & .578 & .645 & .622 & .636 & .709 & .728 & .703 & .707 & .741 & .741 & .752 & .770 & .863 & .848 & .842 & .858 & .872 & .900 & .934 & .937& \textbf{.940}\\
    & $F_{\beta}\uparrow$     & .345 & .502 & .483 & .412 & .585 & .513 & .576 & .542 & .523 & .618 & .604 & .713 & .756 & .735 & .765 & .790 & .824 & .873 & .919 & \textbf{.929}& .928\\
    & $E_{\xi}\uparrow$       & .477 & .572 & .566 & .503 & .748 & .613 & .650 & .631 & .621 & .706 & .684 & .809 & .826 & .825 & .838 & .863 & .888 & .933 & .967 & \textbf{.975}& \textbf{.975}\\
    & $\mathcal{M}\downarrow$ & .114 & .100 & .299 & .168 & .115 & .169 & .208 & .111 & .122 & .090 & .093 & .068 & .055 & .065 & .049 & .046 & .038 & .030 & .019 & \textbf{.016}& \textbf{.016}\\ \hline
  \multirow{4}{*}{\textit{NLPR}\cite{peng2014rgbd}}
    & $S_{\alpha}\uparrow$    & .630 & .632 & .572 & .655 & .727 & .673 & .762 & .724 & .805 & .756 & .806 & .799 & .860 & .856 & .874 & .886 & .888 & .899 & \textbf{.920} & .917& .919\\
    & $F_{\beta}\uparrow$     & .427 & .421 & .430 & .451 & .609 & .429 & .636 & .542 & .649 & .624 & .664 & .755 & .740 & .737 & .802 & .819 & .840 & .865 & .891 & \textbf{.893}& .891\\
    & $E_{\xi}\uparrow$       & .560 & .567 & .542 & .571 & .782 & .579 & .719 & .684 & .745 & .742 & .757 & .851 & .840 & .841 & .887 & .902 & .918 & .940 & .951 & \textbf{.952}& \textbf{.852}\\
    & $\mathcal{M}\downarrow$ & .108 & .108 & .312 & .146 & .112 & .179 & .081& .117 & .095 & .091 & .079 & .058 & .056 & .059 & .044 & .041 & .036 & .031 & .025 & .025& \textbf{.024}\\ \hline
  \multirow{4}{*}{\textit{LFSD}\cite{li2014saliency}}
    & $S_{\alpha}\uparrow$    & .557 & .520 & .722 & .640 & .717 & .734 & .736 & .753 & .700 & .698 & .791 & .738 & .796 & .787 & .794 & .801 & .828 & .847 & .864  & \textbf{.868} & .866\\
    & $F_{\beta}\uparrow$     & .396 & .376 & .612 & .519 & .680 & .566 & .612 & .655 & .521 & .640 & .679 & .736 & .756 & .722 & .761 & .771 & .811 & .845 & .855 & .857& \textbf{.859}\\
    & $E_{\xi}\uparrow$       & .491 & .465 & .638 & .584 & .754 & .625 & .670 & .682 & .588 & .653 & .725 & .796 & .810 & .775 & .818 & .821 & .863 & .893 & .901 & \textbf{.904}& .903\\
    & $\mathcal{M}\downarrow$ & .211 & .218 & .248 & .183 & .167 & .188 & .208 & .155 & .190 & .167 & .138 & .134 & .119 & .132 & .112 & .111 & .088 & .075 & .066 & \textbf{.065}& \textbf{.065}\\ \hline
   \multirow{4}{*}{\textit{SIP} \cite{sip_dataset}}
    & $S_{\alpha}\uparrow$    & .511 & .557 & .616 & .588 & .595 & .732 & .727 & .683 & .717 & .628 & .653 & .720 & .716 & .833 & .842 & .835 & .850 & .806 & .875 & \textbf{.883}& .876\\
    & $F_{\beta}\uparrow$     & .287 & .341 & .496 & .411 & .482 & .542 & .572 & .500 & .568 & .515 & .465 & .702 & .608 & .771 & .814 & .803 & .821 & .811 & .867 & \textbf{.877}& .863\\
    & $E_{\xi}\uparrow$       & .437 & .455 & .564 & .511 & .683 & .614 & .651 & .598 & .645 & .592 & .565 & .793 & .704 & .845 & .878 & .870 & .893 & .844 & .914 & \textbf{.927}& .921\\
    & $\mathcal{M}\downarrow$ & .184 & .192 & .298 & .173 & .224 & .172 & .200 & .186 & .167 & .164 & .185 & .118 & .139 & .086 & .071 & .075 & .064 & .085 & .051 & \textbf{.045}& .049\\
  \hline
  \end{tabular}
\end{table*}

\begin{figure*}[!htp]
  \begin{center}
  {\includegraphics[width=0.97\linewidth]{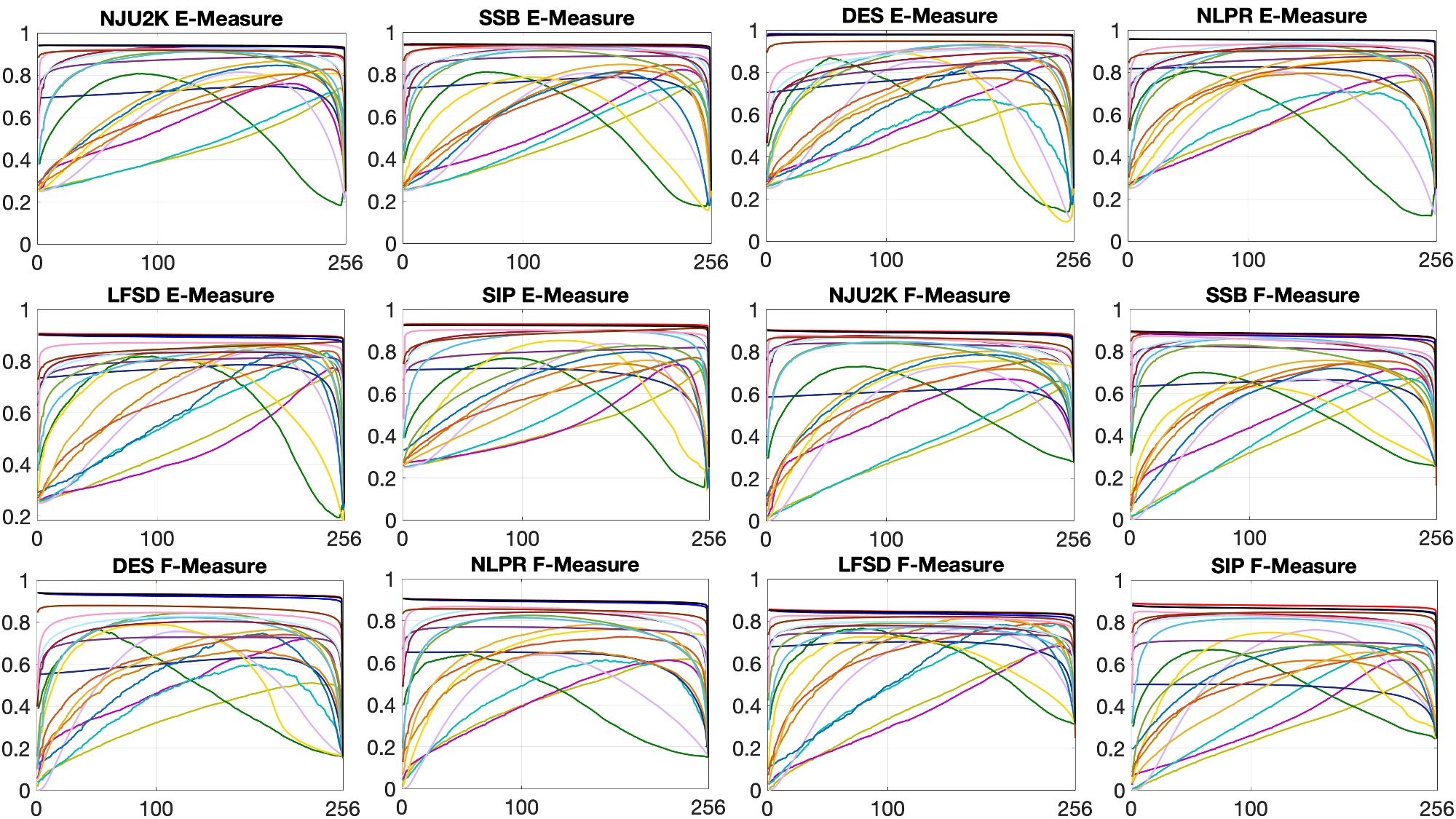}}
\vspace{5mm}
  {\includegraphics[width=0.85\linewidth]{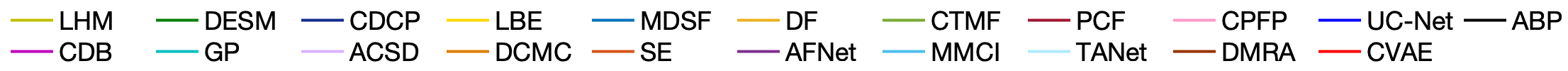}}
  \end{center}
  \vspace{-3mm}
  \caption{E-measure and F-measure curves on six testing datasets (NJU2K, SSB, DES, NLPR, LFSD and SIP). Best viewed on screen.} 
  \label{fig:E_F_measure_show}
\end{figure*}

\begin{table*}[t!]
  \centering
  \small
  \renewcommand{\arraystretch}{1.2}
  \renewcommand{\tabcolsep}{1.7mm}
  \caption{
 The code type and inference time of existing approaches.
  M = Matlab. 
  Pt = PyTorch. Tf = Tensorflow. 
  }\label{tab:ModelSummary}
  \begin{tabular}{l|c|c|c|c|c|c|c|c|c}
  \hline
    Method    & LHM~\cite{peng2014rgbd} & CDB~\cite{liang2018stereoscopic} & DESM~\cite{cheng2014depth} & GP~\cite{ren2015exploiting}
              & CDCP~\cite{zhu2017innovative} & ACSD~\cite{ju2014depth}  & LBE~\cite{feng2016local} &DCMC~\cite{cong2016saliency}& MDSF~\cite{song2017depth}\\
  \hline
    Time (s)  & 2.13  & 0.60 & 7.79 & 12.98 & 60.00 & 0.72 & 3.11 & 1.20 & 60.00 \\
  \hline
    Code Type & M       & M      & M    & M\&C++ & M\&C++ & C++      & M\&C++ & M & C++  \\
  \hline
    Method    &SE~\cite{guo2016salient} & DF~\cite{qu2017rgbd}  & AFNet~\cite{wang2019adaptive} & CTMF~\cite{han2017cnns}
              & MMCI~\cite{chen2019multi} &PCF~\cite{chen2018progressively}& CPFP~\cite{zhao2019Contrast}& Our\_ABP & Our\_CVAE\\
  \hline
    Time (s)  & 1.57 & 10.36 & 0.03 & 0.63 & 0.05 & 0.06 & 0.17 & 0.05 &0.06 \\
  \hline
    Code Type & M\&C++ & M\&C++ & Tf & Caffe & Caffe & Caffe & Caffe & Pt & Pt\\
  \hline
  \end{tabular}
\end{table*}

\subsection{Setup}
\noindent\textbf{Datasets:}
We perform experiments on six datasets including five widely used RGB-D saliency detection datasets (namely NJU2K \cite{NJU2000}, NLPR \cite{peng2014rgbd}, SSB \cite{niu2012leveraging}, LFSD \cite{li2014saliency}, DES \cite{cheng2014depth}) and one newly released dataset (SIP \cite{sip_dataset}). 

\noindent\textbf{Competing Methods:}
We compare our method with 18 algorithms, including ten handcrafted conventional methods and eight deep RGB-D saliency detection models. 


\begin{figure*}[!t]
  \begin{center}
  \begin{tabular}{ c@{ }}
  {\includegraphics[width=0.99\linewidth]{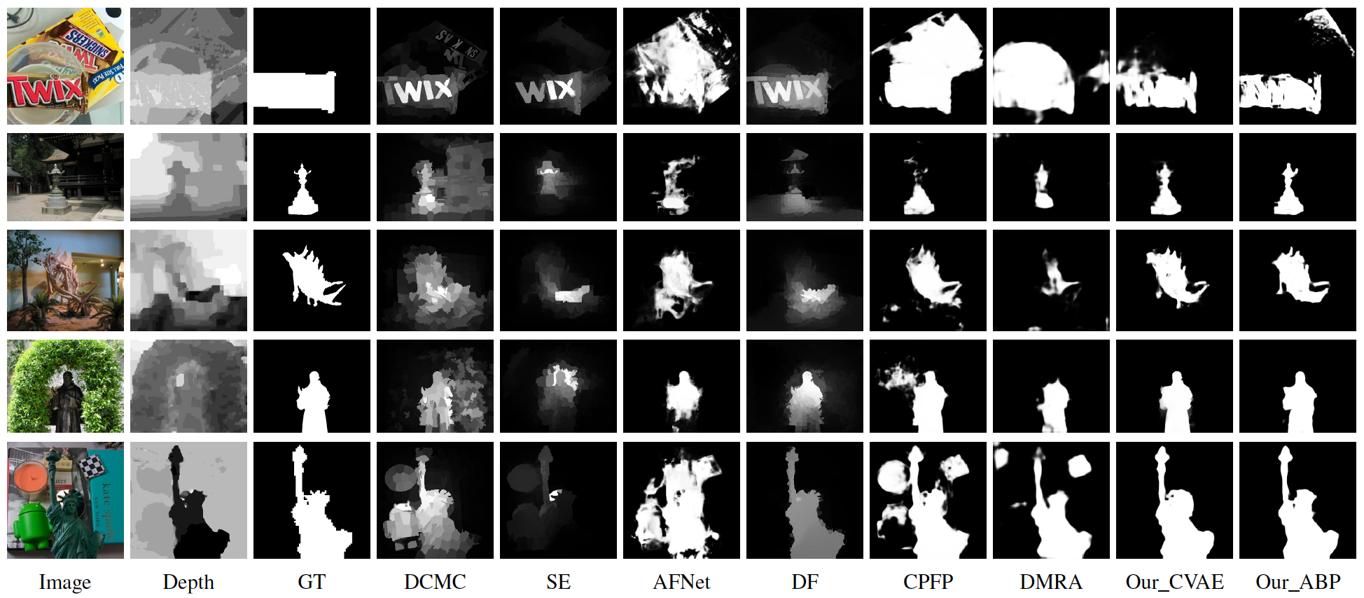}}\\
  \end{tabular}
  \end{center}
  \vspace{-2mm}
\caption{Visual comparison of predictions of our methods and competing methods.
Note that, our final prediction is generated with the proposed \enquote{Saliency Consencus Module} (see Section \ref{determnistic_inference_sec}).
}
  \label{fig:saliency_compare}
\end{figure*}

\begin{table}[t!]
  \centering
  \scriptsize
  \renewcommand{\arraystretch}{1.2}
  \renewcommand{\tabcolsep}{0.4mm}
  \caption{Performance of competing RGB saliency detection models and ours on RGBD saliency datasets, where depth data is not used while testing using the RGB saliency models. We adopt mean $F_{\beta}$ and mean $E_{\xi}$.
  }\label{tab:performance_rgb_models}
  \begin{tabular}{lr|cccccccc|c}
  \hline
    & Metric &
   AFBNet  & NLDF & PiCANet  & RAS     & DGRL & CPD & SCRN & F3Net & CAVE \\
   && \cite{AFNet_Sal}&\cite{Luo_2017_CVPR}&\cite{liu2018picanet}&\cite{ras_reverse}&\cite{wang2018detect}&\cite{CPD_Sal}&\cite{SCRN_iccv} &\cite{F3Net_aaai2020}& Ours\\
  \hline
  \multirow{4}{*}{\textit{NJU2K}\cite{ju2014depth}}
    & $S_{\alpha}\uparrow$    & .862 & .813 & .864 & .754  & .767 & .875 & .879 & .861 & \textbf{.902}  \\
    & $F_{\beta}\uparrow$       & .835 & .783 & .818 & .744  & .716 & .852 & .863 & .837 & \textbf{.893}    \\
    & $E_{\xi}\uparrow$         & .888 & .848 & .869 & .800  & .804 & .903 & .912 & .890 & \textbf{.937}   \\
    & $\mathcal{M}\downarrow$   & .064 & .091 & .072 & .115  & .107 & .056 & .052 & .061 & \textbf{.039}  \\ \hline
  \multirow{4}{*}{\textit{SSB}\cite{niu2012leveraging}}
    & $S_{\alpha}\uparrow$      & .893 & .859 & .896 & .828  & .824 & .902 & .902 & .891 & \textbf{.898}   \\
    & $F_{\beta}\uparrow$      & .865 & .831 & .844 & .820  & .781 & .880 & .881 & .868 & \textbf{.878}    \\
    & $E_{\xi}\uparrow$         & .918 & .893 & .899 & .871  & .865 & .928 & .928 & .921 & \textbf{.935}   \\
    & $\mathcal{M}\downarrow$   & .045 & .062 & .053 & .076  & .073 & .040 & .041 & .043 & \textbf{.039}   \\ \hline
  \multirow{4}{*}{\textit{DES}\cite{cheng2014depth}} 
    & $S_{\alpha}\uparrow$      & .879 & .828 & .883 & .806 & .833 & .894 & .907 & .880 & \textbf{.937}    \\
    & $F_{\beta}\uparrow$       & .845 & .758 & .822 & .762 & .753 & .870 & .885 & .845 & \textbf{.929}   \\
    & $E_{\xi}\uparrow$         & .893 & .831 & .872 & .823 & .849 & .907 & .927 & .892 & \textbf{.975}  \\
    & $\mathcal{M}\downarrow$   & .035 & .058 & .039 & .060 & .054 & .029 & .026 & .030 & \textbf{.016}  \\ \hline
  \multirow{4}{*}{\textit{NLPR}\cite{peng2014rgbd}}
    & $S_{\alpha}\uparrow$      & .881 & .847 & .876 & .853 & .840 & .893 & .894 & .884 & \textbf{.917}    \\
    & $F_{\beta}\uparrow$      & .816 & .782 & .789 & .810 & .767 & .844 & .846 & .838 & \textbf{.893}   \\
    & $E_{\xi}\uparrow$        & .896 & .876 & .870 & .888 & .873 & .914 & .920 & .912 & \textbf{.952}  \\
    & $\mathcal{M}\downarrow$  & .042 & .052 & .051 & .049 & .053 & .034 & .036 & .035 & \textbf{.025}  \\ \hline
  \multirow{4}{*}{\textit{LFSD}\cite{li2014saliency}}
    & $S_{\alpha}\uparrow$     & .817 & .777 & .827 & .673 & .782 & .836 & .827 & .835 & \textbf{.868}   \\
    & $F_{\beta}\uparrow$      & .784 & .756 & .778 & .672 & .759 & .811 & .800 & .810 & \textbf{.857}  \\
    & $E_{\xi}\uparrow$        & .838 & .806 & .825 & .727 & .817 & .856 & .847 & .857 & \textbf{.904}  \\
    & $\mathcal{M}\downarrow$  & .094 & .121 & .103 & .162 & .117 & .088 & .088 & .089 & \textbf{.065}  \\ \hline
   \multirow{4}{*}{\textit{SIP}\cite{sip_dataset}}
    & $S_{\alpha}\uparrow$     & .876 & .795 & .851 & .718 & .682 & .870 & .866 & .866 & \textbf{.883}  \\
    & $F_{\beta}\uparrow$      & .847 & .752 & .806 & .696 & .606 & .859 & .861 & .850 & \textbf{.877}  \\
    & $E_{\xi}\uparrow$        & .911 & .840 & .866 & .766 & .744 & .910 & .903 & .905 & \textbf{.927}  \\
    & $\mathcal{M}\downarrow$  & .055 & .100 & .073 & .121 & .138 & .053 & .057 & .055 & \textbf{.045}  \\
  \hline
  \end{tabular}
\end{table}

\noindent\textbf{Evaluation Metrics:}
Four evaluation metrics are used to evaluate the deterministic predictions, including two widely used: 1) Mean Absolute Error (MAE $\mathcal{M}$); 2) mean F-measure ($F_{\beta}$) and two recently proposed: 3) Structure measure (S-measure, $S_{\alpha}$) \cite{fan2017structure} and 4) mean Enhanced alignment measure (E-measure, $E_{\xi}$) \cite{Fan2018Enhanced}.
\begin{itemize}
    \item \textbf{MAE $\mathcal{M}$:} The MAE estimates the approximation degree between the saliency map $Sal$ and the ground-truth $G$.
    It provides a direct estimate of conformity between estimated and GT map. MAE is defined as:
\begin{equation}
    \begin{aligned}
    \text{MAE} = \frac{1}{N}|Sal-G|,
    \end{aligned}
\end{equation}
where $N$ is the total number of pixels. 

\item \textbf{S-measure $S_{\alpha}$:} Both MAE and F-measure metrics ignore the important structure information evaluation, whereas behavioral vision studies have shown that the human visual system is highly sensitive to structures in scenes~\cite{fan2017structure}.
Thus, we additionally include the structure measure (S-measure~\cite{fan2017structure}).
The S-measure combines the region-aware ($S_r$) and
object-aware ($S_o$) structural similarity as their final structure metric:
\begin{equation}
\label{equ:S-measure}
S_{\alpha} = \alpha*S_o+(1-\alpha)*S_r,
\end{equation}
where $\alpha\!\in\![0,1]$ is a balance parameter and set to 0.5 as default.

\item \textbf{E-measure $E_{\xi}$:} E-measure is the recent proposed Enhanced alignment measure~\cite{Fan2018Enhanced} in the binary map evaluation field. This measure is based on cognitive vision studies, which combines local pixel values with the image-level mean value in one term, jointly capturing image-level statistics and local pixel matching information. Here, we introduce it to provide a more comprehensive evaluation.

\item \textbf{F-measure $F_{\beta}$:} It is essentially a region based similarity metric.
We provide the mean F-measure using varying 255 fixed (0-255) thresholds as shown in Fig. \ref{fig:E_F_measure_show}.
\end{itemize}

\noindent\textbf{Implementation Details:}
We train our model using PyTorch, and initialized the encoder of the
\enquote{Generator Model} with ResNet50 \cite{ResHe2015} parameters pre-trained on ImageNet. Inside the \enquote{DASPP} module of the \enquote{Generator Model} in Fig. \ref{fig:saliency_feature_net}, we use four different scales of dilation rate: {6, 12, 18, 24} same as \cite{denseaspp}, and set all intermediate channel size as $M=32$. For both inference models, we set the dimension of the latent variable as $K=3$. Weights of new layers are initialized with $\mathcal{N}(0,0.01)$, and bias is set as constant. We use the Adam method with momentum 0.9 and decrease the learning rate 10\% after 80\% of the maximum epoch. The base learning rate is initialized as 5e-5. The whole training takes around 9 hours with training batch size 5, and maximum epoch 100 on a PC with an NVIDIA GeForce RTX GPU. For input image size $352\times352$, the inference time of our CVAE model and ABP model
are 0.06s and 0.05s on average respectively. 
\subsection{Comparison to State-of-the-art Methods
}

\noindent\textbf{Quantitative Comparison:}
We report the performance of our method (with both inference models) and competing methods in Table \ref{tab:BenchmarkResults}, where \enquote{CVAE} is our framework with CVAE as inference model, and \enquote{ABP} represents the model that updates latent variable $z$ with alternating back-propagation. Results in Table \ref{tab:BenchmarkResults} demonstrate the benefits of both CVAE and ABP which consistently achieve the best performance on all datasets. Specifically, on SSB \cite{niu2012leveraging} and SIP \cite{sip_dataset},
our method achieves around a 2.5\% S-measure, E-measure and F-measure performance boost and a decrease in MAE by 1.5\% compared with the \enquote{Deep Models} in Table \ref{tab:BenchmarkResults}.
Moreover, compared with our preliminary version \enquote{UC-Net} \cite{jing2020uc}, we observe improved performance, which indicates the effectiveness of the proposed structure.
We also show E-measure and F-measure curves of competing methods and ours in Fig. \ref{fig:E_F_measure_show}. 
We observe that our method produces not only stable E-measure and F-measure but also the best performance.\\
To further evaluate the proposed method, we compute performance of eight cutting-edge RGB saliency detection models on the RGB-D testing dataset\footnote{The RGB saliency models are trained on RGB saliency training set, and testing on RGB-D testing set, where the depth is not used.} and compared with our \enquote{CVAE} based model. The results are shown in Table \ref{tab:performance_rgb_models}, which further illustrates the superior performance of the proposed framework.

\noindent\textbf{Qualitative Comparisons:}
In Fig. \ref{fig:saliency_compare}, we show five examples comparing our method with six RGB-D saliency detection models.
Salient objects in these images can be large (fifth row), small (second row) or in complex backgrounds (first, third, fourth and fifth rows). Especially for the example in the first row, the background is complex, 
part of the background shares similar color and texture as the salient foreground. Most of those competing methods (AFNet\cite{wang2019adaptive}, CPFP\cite{zhao2019Contrast} and DMRA\cite{dmra_iccv19}) failed to correctly segment the precise salient foreground, while our approach achieves better salient object detection with each of the proposed two inference models. For the image in the last row, there exists an object (\ie, green toy) that strongly stands out from its background, while the depth map can to some extent decrease
the salience of
such high-contrast region. All of the competing methods (DCMC\cite{cong2016saliency}, SE\cite{guo2016salient}, AFNet\cite{wang2019adaptive}, CPFP\cite{zhao2019Contrast} in particular)
falsely detect part of the background region as being salient, whereas our accurate predictions further indicate the effectiveness of our solutions.
With all the results in Fig. \ref{fig:saliency_compare}, we can
see evidence of
the superiority of our approach.

\begin{table*}[t!]
  \centering
  \footnotesize
  \renewcommand{\arraystretch}{1.1}
  \renewcommand{\tabcolsep}{0.38mm}
  \caption{Evaluation of the effect of different components in our models, and alternative structures. We present mean $F_{\beta}$ and mean $E_{\xi}$.
  }
  \begin{tabular}{l|cccc|cccc|cccc|cccc|cccc|cccc}
  \hline
  &\multicolumn{4}{c|}{NJU2K}&\multicolumn{4}{c|}{SSB}&\multicolumn{4}{c|}{DES}&\multicolumn{4}{c|}{NLPR}&\multicolumn{4}{c|}{LFSD}&\multicolumn{4}{c}{SIP} \\
    Method & $S_{\alpha}\uparrow$&$F_{\beta}\uparrow$&$E_{\xi}\uparrow$&$\mathcal{M}\downarrow$& $S_{\alpha}\uparrow$&$F_{\beta}\uparrow$&$E_{\xi}\uparrow$&$\mathcal{M}\downarrow$& $S_{\alpha}\uparrow$&$F_{\beta}\uparrow$&$E_{\xi}\uparrow$&$\mathcal{M}\downarrow$& $S_{\alpha}\uparrow$&$F_{\beta}\uparrow$&$E_{\xi}\uparrow$&$\mathcal{M}\downarrow$& $S_{\alpha}\uparrow$&$F_{\beta}\uparrow$&$E_{\xi}\uparrow$&$\mathcal{M}\downarrow$ & $S_{\alpha}\uparrow$&$F_{\beta}\uparrow$&$E_{\xi}\uparrow$&$\mathcal{M}\downarrow$\\
  \hline
   Middle & .897 & .888 & .933 & .042 & .895 & .880 & .934 & .041 & .931 & .920 & .968 & .018 & .916 & .887 & .950 & .026 & .854 & .843 & .888 & .073 & .873 & .863 & .914 & .048   \\
   Late & .890 & .875 & .929 & .046 & .891 & .866 & .931 & .042 & .929 & .909 & .970 & .020 & .907 & .877 & .947 & .028 & .839 & .828 & .887 & .076 & .870 & .853 & .916 & .051   \\
   AveP & .900 & .892 & .936 & .040 & .897 & .877 & .934 & .040 & .935 & .924 & .970 & .017 & .914 & .890 & .951 & .025 & .857 & .842 & .899 & .067 & .880 & .876 & .926 & .046   \\
   AveZ & .901 & .890 & .927 & .040 & .892 & .875 & .930 & .040 & .929 & .921 & .971 & .018 & .914 & .884 & .950 & .026 & .855 & .843 & .892 & .068 & .880 & .874 & .926 & .046   \\ 
   GSNN & .900 & .887 & .935 & .040 & .894 & .873 & .930 & .041 & .931 & .919 & .971 & .018 & .913 & .885 & .949 & .026 & .852 & .834 & .894 & .070 & .871 & .864 & .916 & .051   \\
   CVAE\_S & .900 & .890 & .932 & .040 & .894 & .876 & .931 & .041 & .936 & .927 & .974 & .016 & .914 & .891 & .949 & .026 & .856 & .843 & .897 & .068 & .877 & .867 & .920 & .048  \\
   NoS & .893 & .881 & .933 & .042 & .885 & .876 & .930 & .044 & .931 & .921 & .966 & .017 & .914 & .878 & .950 & .027 & .853 & .845 & .898 & .069 & .882 & .868 & .924 & .047   \\
   CE & .900 & .891 & .936 & .041 & .894 & .876 & .930 & .040 & .935 & .921 & .970 & .018 & .913 & .891 & .950 & .025 & .851 & .833 & .887 & .075 & .876 & .856 & .916 & .051   \\
   HHA & .897 & .886 & .934 & .042 & .902 & .882 & .937 & .038 & .930 & .917 & .970 & .019 & \textbf{.919} & .892 & .950 & .024 & .850 & .834 & .888 & .074 & .870 & .856 & .915 & .052   \\
   w/o KLA & .900 & .890 & .932 & .041 & .893 & .870 & .931 & .040 & .932 & .923 & .972 & .017 & .913 & .887 & .948 & .027 & .854 & .841 & .893 & .069 & .881 & .872 & .923 & .046   \\ \hline
    Our\_CAVE    & \textbf{.902} & \textbf{.893} & \textbf{.937} & \textbf{.039} &         .898  & .878 & .935 & .039 & .937 & \textbf{.929} & \textbf{.975} & \textbf{.016} &         .917  & \textbf{.893} & \textbf{.952} & .025 & \textbf{.868} & .857 & \textbf{.904} & \textbf{.065} & \textbf{.883} & \textbf{.877} & \textbf{.927} & \textbf{.045}   \\
   Our\_ABP    & .900 & .889 & \textbf{.937} & \textbf{.039} &         \textbf{.904}  & \textbf{.886} & \textbf{.939} & \textbf{.037} & \textbf{.940} & .928 & \textbf{.975} & \textbf{.016} &        \textbf{.919}  & .891 & \textbf{.852} & \textbf{.024} & .866 & \textbf{.859} & .903 & \textbf{.065} & .876 & .863 & .921 & .049   \\
   \hline
  \end{tabular}
  \label{tab:model_analysis_models}
\end{table*}

\begin{figure*}[!t]
   \begin{center}
   \begin{tabular}{ c@{ }}
  {\includegraphics[width=0.99\linewidth]{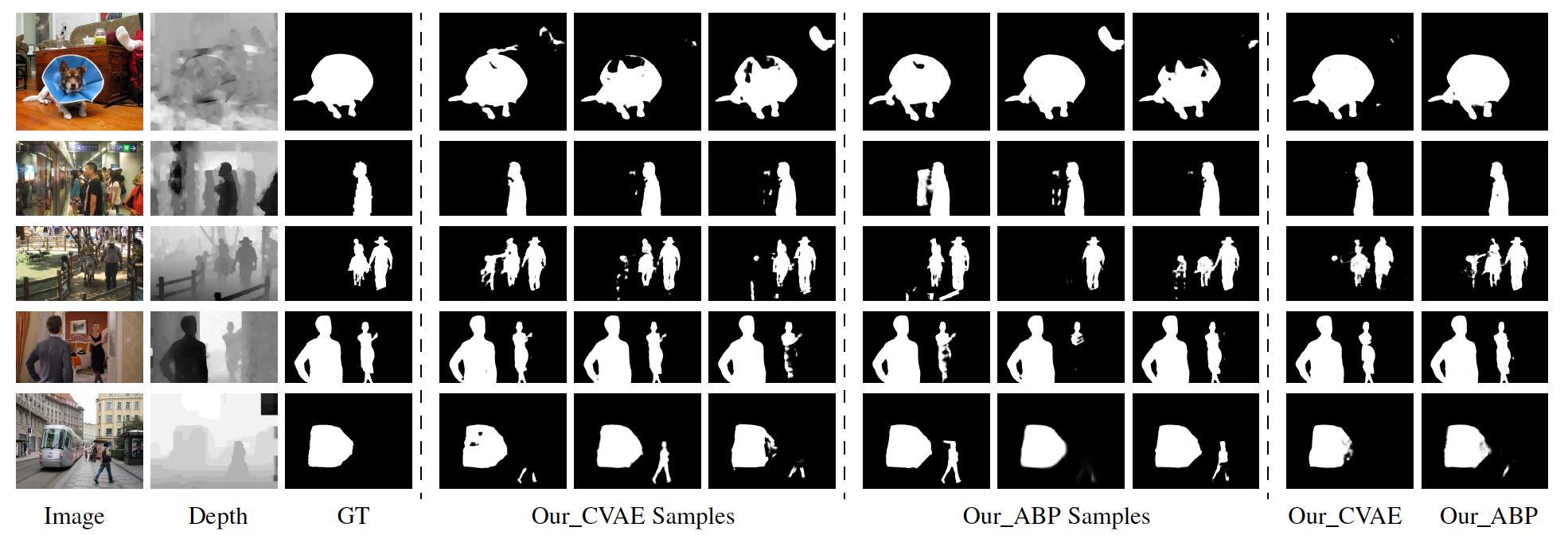}} \\
   \end{tabular}
   \end{center}
  \vspace{-2mm}
\caption{Structured outputs generation,
where \enquote{Our\_CVAE Samples} and \enquote{Our\_CVAE} are samples and the deterministic prediction respectively.
}
   \label{fig:our_samples}
\end{figure*}

\begin{figure}[!htp]
   \begin{center}
   {\includegraphics[width=\linewidth]{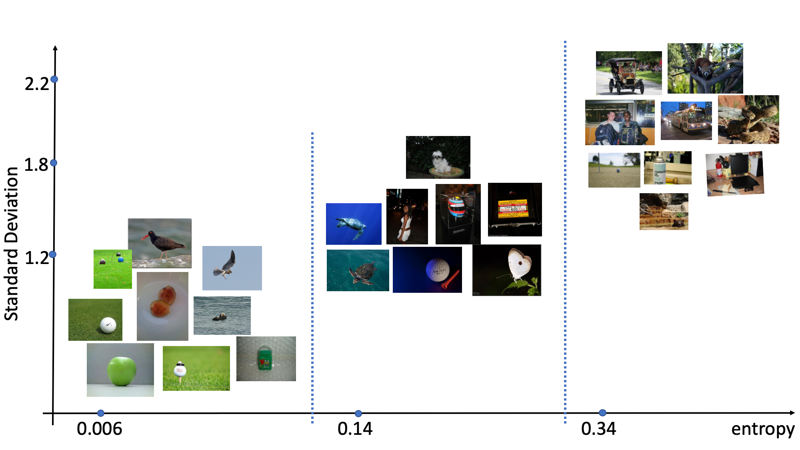}} 
   \end{center}
   \vspace{-5pt}
   \caption{Image distribution by analysing entropy and standard deviation.}
   \label{fig:diversity_entropy_connection}
\end{figure}

\noindent\textbf{Probabilistic Distribution Evaluation:} As a probabilistic network, our models can produce a distribution of plausible saliency maps instead of a single, deterministic prediction for each input image. We argue that, for images with simple background, consistent predictions should be produced, whereas for complex images with cluttered background, we expect our model to capture the uncertainty in the saliency maps, and thus can generate diverse predictions.
To evaluate performance of our model, following the active learning pipeline \cite{Settles10activelearning}, we first generate $B=100$
easy and difficult samples. To achieve this, we first adopt three different conventional saliency models (RBD \cite{Background-Detection:CVPR-2014}, MR \cite{Manifold-Ranking:CVPR-2013} and GS \cite{GS_Sal}, which rank among the top six conventional handcrafted feature based RGB saliency models \cite{borji2015salient}), and define them as $f1$, $f2$ and $f3$ respectively.
Given image $X_i$\footnote{We use the RGB data only.} in training dataset $D$, we compute its corresponding saliency map $f1(X_i)$, $f2(X_i)$ and $f3(X_i)$. We choose entropy as measure for image complexity. Then, we define mean saliency map of $X_i$ as $P_i=(f1(X_i)+f2(X_i)+f3(X_i))/3$. We define the
complexity of the
image as task driven (for saliency detection). Then given a
ground-truth saliency map $Y_i$ and mean saliency map $P_i$, we define foreground entropy as: $-P_i\log P_i$. 

\begin{figure*}[t!]
   \begin{center}
   \begin{tabular}{ c@{ } c@{ } c@{ }}
   {\includegraphics[width=0.23\linewidth]{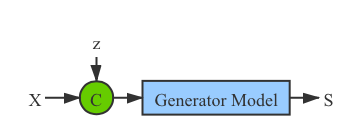}} &
   {\includegraphics[width=0.36\linewidth]{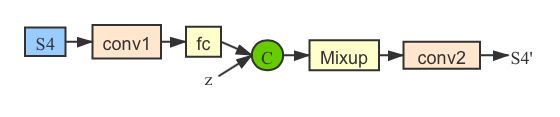}} & {\includegraphics[width=0.36\linewidth]{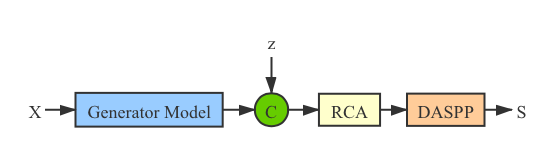}} \\
    \footnotesize{(a) Early fusion model} &\footnotesize{(b) Middle fusion model} &
    \footnotesize{(c) Late fusion model} \\
   \end{tabular}
   \end{center}
   \caption{Detail network structures of different fusion schemes: the early fusion model (a), the middle fusion model (b) and the late fusion model (c).
   }
   \label{fig:middle_late_fusion_model}
\end{figure*}

We then define mean entropy as a complexity measure, and choose $B$
images with the smallest entropy as the easy samples and $B$
images with the largest entropy as the difficult samples (with $B=100$).
We sample $Sn=5$ times from the prior distribution and compute the variance of each group.
Specifically, for image pair $X_i$,
with 
$Sn$ iterations
of sampling, we obtain its prediction $\{S_i^j\}_{j=1}^{Sn}$. We compute the similarity of these $Sn$ different predictions, and treat it as prediction diversity evaluation.
We show entropy and standard deviation of images in Fig. \ref{fig:diversity_entropy_connection}.

\noindent\textbf{Inference Time\footnote{Conventional handcrafted-feature based methods are implemented on CPU, and deep RGB-D saliency prediction models are based on GPU, thus we report CPU time for the former and GPU time for the later.} Comparison:}
We summarize basic information of
competing methods in Table \ref{tab:ModelSummary} for clear comparison, including their code type and inference time.
Table \ref{tab:ModelSummary} shows that the inference time\footnote{The inference time we report represents prediction with one random sampling from the PriorNet.} of our method is
comparable with competing methods, which further illustrates that our model can achieve probabilistic predictions with no inference time sacrificed.


\begin{figure}[!htp]
   \begin{center}
   {\includegraphics[width=0.99\linewidth]{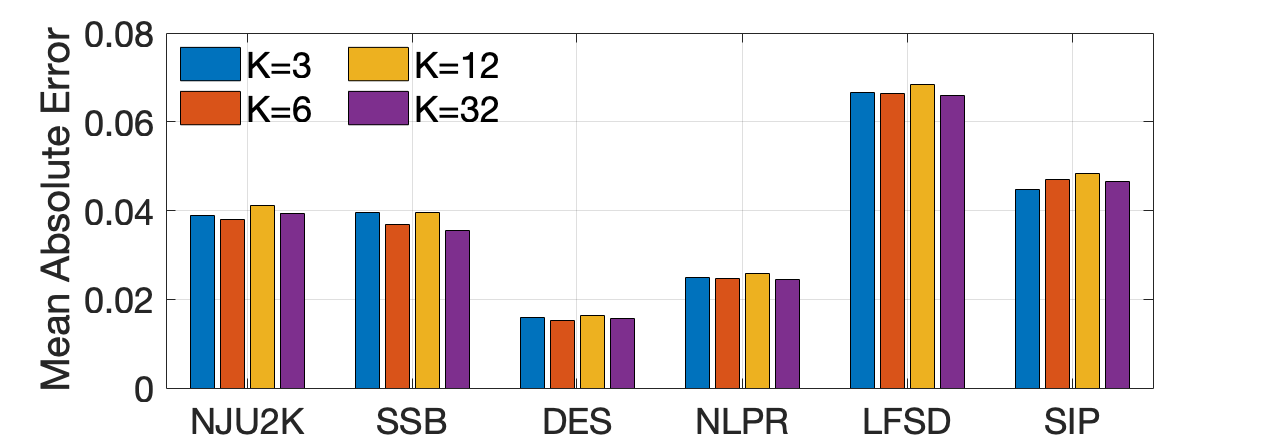}} 
   \end{center}
   \caption{Dimension analysis of the latent variable.}
   \label{fig:influence_of_scale_of_latent_space}
\end{figure}

\subsection{Structured Output Generation}
As a generative network, we introduce a latent variable $z$ modeling uncertainty of human annotation. We further show examples of our model generating structured outputs as shown in Fig. \ref{fig:our_samples}. The \enquote{Our\_CVAE Samples}
in Fig. \ref{fig:our_samples} represents three random samples of our method with the CVAE inference model, and \enquote{Our\_ABP Samples} are samples with the ABP strategy. \enquote{Our\_CVAE} and \enquote{Our\_ABP} are the deterministic predictions of our frameworks with the above two inference models obtained via our \enquote{Saliency Consensus Module}. Fig. \ref{fig:our_samples} shows that both the two inference models can produce reasonable stochastic predictions, and the final deterministic prediction after the \enquote{Saliency Consensus Module} (\enquote{Our\_CVAE} and \enquote{Our\_ABP}) is consistent with the provided GT, which verifies effectiveness of both our latent variable and the \enquote{Saliency Consensus Module}.

\begin{figure*}[!htp]
  \begin{center}
  {\includegraphics[width=0.31\linewidth]{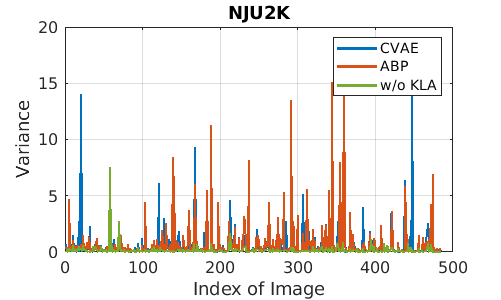}}
  {\includegraphics[width=0.31\linewidth]{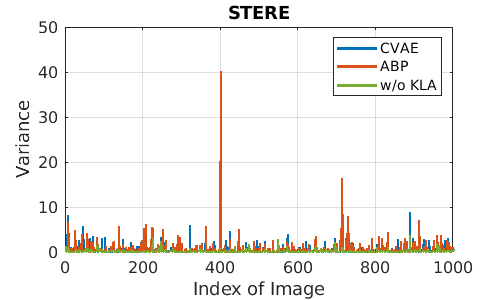}}
  {\includegraphics[width=0.31\linewidth]{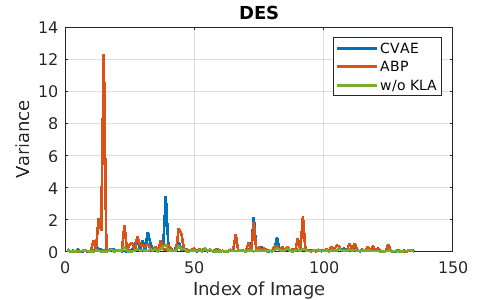}}
  {\includegraphics[width=0.31\linewidth]{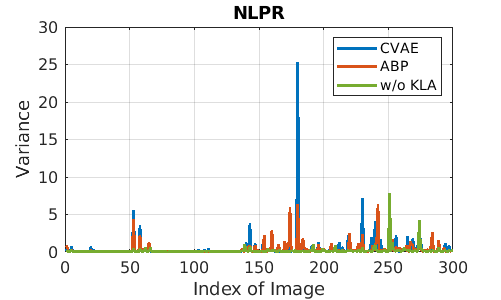}}
   {\includegraphics[width=0.31\linewidth]{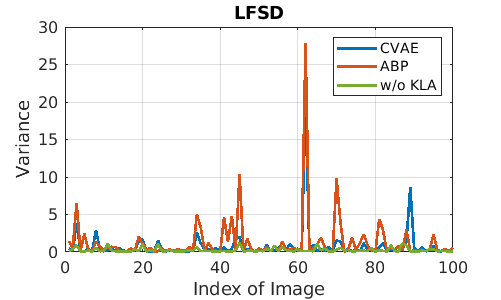}}
  {\includegraphics[width=0.31\linewidth]{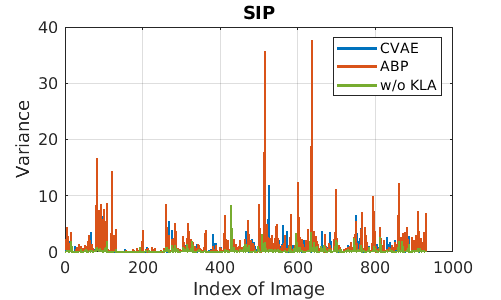}}
  \end{center}
  \vspace{-3mm}
  \caption{Mean variance of multiple predictions using our CAVE-based model (\enquote{CVAE}), ABP-based model (\enquote{ABP}), and the CAVE-based model without KL annealing term (\enquote{w/o KLA}). Best viewed on screen.} 
  \label{fig:variance_maps}
\end{figure*}

\subsection{Ablation Studies
}
We further analyse the proposed framework in this section, including the generative network related strategies, the loss functions, the alternative depth data (HHA \cite{gupta2014learning} in particular), and the solution to prevent network from posterior collapse.
We show the performance in Table \ref{tab:model_analysis_models}. Note that unless otherwise stated, we use the CVAE-based inference model in the following experiments.

\noindent\textbf{Different Fusion Schemes}: The latent variable $z$ can be fused to the network in three different ways: early fusion (in the input layer), middle fusion (in bottleneck network), or late fusion (before the output layer). We propose an early fusion model as shown in Fig. \ref{fig:middle_late_fusion_model} (a).
We further design a middle fusion models and a late fusion model
as shown in Fig. \ref{fig:middle_late_fusion_model} (b) and (c) respectively. The performance of each model is shown in Table \ref{tab:model_analysis_models} \enquote{Middle} and \enquote{Late}.
For the middle fusion model, last convolutional layer of the fourth group (\eg, S4) of the backbone network is fed to a $1\times1$ convolutional layer to obtain a $M=32$ dimensional feature map, which is then map to a $K$ (dimension of the latent variable $z$) dimensional feature vector with a fully connected layer (\enquote{fc}). To avoid posterior collapse \cite{Lagging_Inference_Networks}, inspired by~\cite{aliakbarian2019learning}, 
we mix (\enquote{Mixup}) the feature vector and $z$ channel-wise; thus, the network cannot distinguish between features of the deterministic branch and the probabilistic branch. We then expand the mixed feature vector in the spatial dimension, and feed it to another $1\times1$ convolutional layer to achieve feature map S4' of the same dimension as S4, and replace S4 with S4' in Fig. \ref{fig:saliency_feature_net}. For the late fusion model, the \enquote{Generator Model} represents the generator model in Fig. \ref{fig:saliency_feature_net} before the last \enquote{RCA} module. We expand $z$ in spatial dimension and concatenate
it with the deterministic feature. We also perform \enquote{Mixup} here similar to the middle fusion model. We then feed the mixed feature map to one \enquote{RCA} module and \enquote{DASPP} model to achieve prediction $S$.
We observe slightly worse performance of the middle fusion model (\enquote{Middle}) and late fusion model (\enquote{Late}). The main reason is that strong non-linear representation can be obtained when the latent variable is fed to the beginning of the network, which is also consistent with the result that \enquote{Middle} is better than \enquote{Late}.


\noindent\textbf{Analysing the Effect of the Dimension of $z$}: The scale of $z$ may influence both network performance and diversity of predictions. In this paper, we set dimension of $z$ to 3. We further carry out experiments with dimension of $z$ in the range of $[3,32]$, and show mean absolution error of our model on six benchmark RGB-D saliency dataset in Fig. \ref{fig:influence_of_scale_of_latent_space}. We observe relatively stable performance for different dimension of $z$. 
The relatively stable
performance regardless of
dimension of $z$ shows that the capacity of the network is large enough to take different degree of stochasticity in the
input. Meanwhile, as there exists only a few quite difficult samples, and lower dimension of $z$ is enough to capture variants of labeling.

\noindent\textbf{Deterministic Prediction Generation:} As introduced in Section \ref{determnistic_inference_sec}, three different solutions can be used to generate a deterministic prediction for performance evaluation, including 1) averaging multiple predictions; 2) averaging multiple latent variables; and 3) the proposed saliency consensus module. We evaluate
performance of other deterministic inference solutions and show performance in Table \ref{tab:model_analysis_models} \enquote{AveP} and \enquote{AveZ}, representing the average-prediction solution and average-$z$ solution respectively. We observe similar performance of \enquote{AveP} and \enquote{AveZ} compared with the proposed saliency consensus module. The similar performance of \enquote{AveP} and \enquote{AveZ} illustrates that both conventional deterministic prediction generation solutions work well for the saliency detection task. The better performance of \enquote{Ours} indicates effectiveness
of the proposed solution.

\noindent\textbf{Effectiveness of Loss Functions:}
Due to the inconsistency of $Q_\phi(z|X,Y)$ and $P_\theta(z|X)$ used in the training and testing stage respectively, the model may behave differently during training and testing. To mitigate the discrepancy in encoding the latent variable, and achieve similar network behavior during training and testing, we introduce Gaussian Stochastic Neural Network (GSNN) and a hybrid loss function as shown in Eq. \ref{hybrid_func}. To test how our network performs with only the CVAE loss in Eq. \ref{CVAE_equation} or GSNN loss in Eq. \ref{gsnn_equ}, we train two extra models and show performance as \enquote{CVAE\_S} and \enquote{GSNN} respectively. We see clear performance decreased with each loss used solely. Meanwhile, although the two models perform worse than the proposed solution, we still observe consistent better performance compared with competing methods. Both the performance drop of \enquote{CVAE\_S} and \enquote{GSNN} compared with \enquote{Ours}, and better performance of \enquote{CVAE\_S} and \enquote{GSNN} compared with competing methods, indicate effectiveness of the proposed generative model for saliency detection.

\noindent\textbf{Smoothness Loss:} We introduce the smoothness loss to our loss function to set constraints on the structure of the prediction. To evaluate the contribution of the smoothness loss, we
remove it from our loss function and show the performance as \enquote{NoS}. The lower performance indicates the effectiveness of the smoothness loss. Moreover, as shown in Eq. \ref{smoothness_loss}, the smoothness loss takes saliency prediction and gray-scale
image as input, which can also be interpreted as a self-supervised regularizer.

\noindent\textbf{Structure-aware Loss $vs.$ Cross-entropy Loss:} Similar to \cite{F3Net_aaai2020}, we use structure-aware loss instead of the widely used cross-entropy loss to penalize prediction along object edges, thus we can achieve structure-preserving saliency prediction. To prove that our model can also works well with basic cross-entropy loss, we designed another model with cross-entropy loss used instead of the structure-aware loss, and show performance as \enquote{CE}. We notice clear decreased performance of \enquote{CE} on \enquote{LFSD} and \enquote{SIP} dataset. For both \enquote{LFSD} and \enquote{SIP} dataset, there exists salient foreground regions that share similar color as the background, which makes the cross-entropy based model ineffective in those scenarios. While the structure-aware loss can penalize prediction with wrong structure information, making it effective for those difficult images.

\noindent\textbf{HHA $vs.$ Depth:}
HHA \cite{gupta2014learning} is a widely used technique that encodes the depth data to three channels: \textbf{h}orizontal disparity, \textbf{h}eight above ground, and the \textbf{a}ngle the pixel’s local surface normal makes with the inferred gravity direction. HHA is widely used in RGB-D dense models \cite{Du_2019_CVPR, han2017cnns} to obtain better feature representation. To test if HHA also works in our scenario, we replace depth with HHA, and performance is shown in \enquote{HHA}. We observe similar performance achieved with HHA instead of the raw depth data. Those models using HHA aim to obtain better depth representation, as the raw depth is not usually in low-quality. The proposed stochastic model introduces randomness to the network, which can also serve as denoising technique to improve robustness of the model, and this is also consistent with the observation in \cite{train_with_noise1995}.

\noindent\textbf{Training without KL Annealing:} As discussed in Section \ref{unique_solution}, we introduce KL annealing strategy to prevent the possible posterior collapse problems of the CVAE-based model. To test contribution of this strategy, we simply remove the KL annealing term, and set weight of the KL loss term in Eq. \ref{CVAE_equation} as 1 from the first epoch. Performance of this experiment is shown as \enquote{w/o KLA}. Although the performance on the six benchmark RGB-D saliency datasets does not show effect of KL annealing clearly (as we generate a deterministic prediction), we observed that it highly affects the diversity of the prediction as shown in Fig. \ref{fig:variance_maps}, which presents the mean variance of multiple predictions on the RGB-D testing sets. Specifically, we perform five iterations of random sampling during testing, and compute variance of those five different predictions. We show mean of the variance maps in Fig. \ref{fig:variance_maps}. Meanwhile, we show the mean variance of our CVAE-based and ABP-based models as \enquote{CVAE} and \enquote{ABP} respectively.
Fig. \ref{fig:variance_maps} clearly shows that both of our proposed solutions can generate more diverse predictions than \enquote{w/o KLA}, leading to larger variance than \enquote{w/o KLA}.



\begin{table*}[t!]
  \centering
  \footnotesize
  \renewcommand{\arraystretch}{1.1}
  \renewcommand{\tabcolsep}{0.38mm}
  \caption{Comparison with the state-of-the-art RGB saliency detection models on six benchmark RGB saliency datasets. We adopt mean $F_{\beta}$ and mean $E_{\xi}$.
  }
  \begin{tabular}{l|cccc|cccc|cccc|cccc|cccc|cccc}
  \hline
  &\multicolumn{4}{c|}{DUTS}&\multicolumn{4}{c|}{ECSSD}&\multicolumn{4}{c|}{DUT}&\multicolumn{4}{c|}{HKU-IS}&\multicolumn{4}{c|}{THUR}&\multicolumn{4}{c}{SOC} \\
    Method & $S_{\alpha}\uparrow$&$F_{\beta}\uparrow$&$E_{\xi}\uparrow$&$\mathcal{M}\downarrow$& $S_{\alpha}\uparrow$&$F_{\beta}\uparrow$&$E_{\xi}\uparrow$&$\mathcal{M}\downarrow$& $S_{\alpha}\uparrow$&$F_{\beta}\uparrow$&$E_{\xi}\uparrow$&$\mathcal{M}\downarrow$& $S_{\alpha}\uparrow$&$F_{\beta}\uparrow$&$E_{\xi}\uparrow$&$\mathcal{M}\downarrow$& $S_{\alpha}\uparrow$&$F_{\beta}\uparrow$&$E_{\xi}\uparrow$&$\mathcal{M}\downarrow$& $S_{\alpha}\uparrow$&$F_{\beta}\uparrow$&$E_{\xi}\uparrow$&$\mathcal{M}\downarrow$ \\
  \hline
   DGRL \cite{wang2018detect} & .846 & .790 & .887 & .051 & .902 & .898 & .934 & .045 & .809 & .726 & .845 & .063 & .897 & .884 & .939 & .037 & .816 & .727 & .838 & .077& - & - & - & -  \\
   PiCAN \cite{liu2018picanet} & .842 & .757 & .853 & .062 & .898 & .872 & .909 & .054 & .817 & .711 & .823 & .072 & .895 & .854 & .910 & .046 & .818 & .710 & .821 & .084  & .801 & .332 & .810 & .133 \\ 
   NLDF \cite{Luo_2017_CVPR} & .816 & .757 & .851 & .065 & .870 & .871 & .896 & .066 & .770 & .683 & .798 & .080 & .879 & .871 & .914 & .048 & .801 & .711 & .827 & .081  & .816 & .319 & .837 & .106 \\ 
   BASN \cite{BASNet_Sal} & .876 & .823 & .896 & .048 & .910 & .913 & .938 & .040 & .836 & .767 & .865 & .057 & .909 & .903 & .943 & .032 & .823 & .737 & .841 & .073  & .841 & .359 & .864 & .092 \\ 
   AFNet \cite{AFNet_Sal} & .867 & .812 & .893 & .046 & .907 & .901 & .929 & .045 & .826 & .743 & .846 & .057 & .905 & .888 & .934 & .036 & .825 & .733 & .840 & .072  & .700 & .062 & .684 & .115 \\ 
   MSNet \cite{MSNet_Sal} & .862 & .792 & .883 & .049 & .905 & .886 & .922 & .048 & .809 & .710 & .831 & .064 & .907 & .878 & .930 & .039 & .819 & .718 & .829 & .079  & - & - & - & - \\ 
   SCRN \cite{SCRN_iccv} & .885 & .833 & .900 & .040 & .920 & .910 & .933 & .041 & .837 & .749 & .847 & .056 & .916 & .894 & .935 & .034 & .845 & .758 & .858 & .066  & .838 & .363 & .859 & .099 \\
   LDF \cite{ldf_sal_cvpr20} & \textbf{.890} & .861 & .925 & \textbf{.034} & .919 & .923 & .943 & .036 & .839 & .770 & .865 & .052 & .920 & .913 & .953 & .028 & .842 & .768 & .863 & \textbf{.064}  & - & - & - & - \\
   \hline
  Ours\_CVAE & .888 & .860 & .927 & \textbf{.034} & \textbf{.921} & \textbf{.926} & \textbf{.947} & \textbf{.035} & .839 & \textbf{.773} & \textbf{.869} & .051 & \textbf{.921} & \textbf{.919} & \textbf{.957} & \textbf{.026} & .848 & .765 & .862 & \textbf{.064} & \textbf{.849} & \textbf{.369} & \textbf{.872} & \textbf{.089}  \\ 
  Ours\_ABP & \textbf{.890} & \textbf{.864} & \textbf{.931} & \textbf{.034} & .915 & .918 & .941 & .037 & \textbf{.843} & .770 & .864 & \textbf{.050} & .917 & .913 & .949 & .027 & \textbf{.849} & \textbf{.773} & \textbf{.869} & .066 & .842 & .365 & .868 & .091  \\ 
  \hline
  \end{tabular}
  \label{tab:benchmark_model_comparison_rgb}
\end{table*}

\subsection{Probabilistic RGB Saliency Detection}
We propose a generative model based RGB-D saliency detection network, and we extend it to RGB saliency detection to test flexibility of the proposed framework, and show performance in Table \ref{tab:benchmark_model_comparison_rgb}. We train our model (\enquote{Ours\_CVAE} and \enquote{Ours\_ABP}) with DUTS training dataset \cite{imagesaliency}, and evaluate performance of our methods and competing methods on six widely-used benchmarks: (1) DUTS testing dataset; (2) ECSSD \cite{Hierarchical:CVPR-2013}; (3) DUT \cite{Manifold-Ranking:CVPR-2013}; (4) HKU-IS \cite{MDF:CVPR-2015}; (5) THUR \cite{THUR} and (6) SOC \cite{fan2018salient_soc}. 
Note that, similar to the RGB-D based framework, we use the same network structure, except that the input image $X$ is RGB data instead of the RGB-D image pair. The consistent better performance of our network (\enquote{Ours\_CVAE} or \enquote{Ours\_ABP}) illustrates flexibility of our model, which can be lead to new benchmark performance for both RGB-D saliency detection and RGB saliency detection.
\section{Conclusion}
Inspired by human uncertainty in ground-truth annotation, we proposed the first uncertainty inspired RGB-D saliency detection model.
Different from existing methods, which generally treat saliency detection as a point estimation problem, we propose to learn the distribution of saliency maps, and proposed a generative learning pipeline to produce stochastic saliency predictions. Meanwhile, we introduce two different inference models: 1) a CVAE-based inference model, where an extra encoder to approximate true posterior distribution of the latent variable $z$; and 2) an ABP-based inference model to sample $z$ directly from its true posterior distribution with gradient based MCMC.
Under our formulation, our model is able to generate multiple predictions, representing uncertainty of human annotation. With the proposed saliency consensus module, we are able to produce accurate saliency prediction following the similar pipeline as the ground-truth annotation generation process. Quantitative and qualitative evaluations on six standard and challenging benchmark RGB-D datasets demonstrated the superiority of our approach in learning the distribution of saliency maps. 

Meanwhile, we thoroughly investigate the generative model and include analysis of both the latent variable, the loss function and the different fusion schemes to introduce $z$ to the network. Furthermore, we extend our solutions to RGB saliency detection. Without changing network structure (we only change the input from RGB-D data to RGB data), we achieve state-of-the-art performance compared with the last RGB saliency models.

Two different inference models are introduced to learn the proposed generative network as shown in Fig. \ref{fig:simplyfied_training_testing} (a). From our experience, both the CVAE-based and ABP-based inference models can lead to diverse saliency predictions as shown in Fig. \ref{fig:variance_maps}. While, as extra encoder used in the CVAE model, it leads to more network parameters than the ABP-based solution. On the other hand, as we update the latent variable by running several steps of Langevin Dynamics based MCMC as shown in Eq. \ref{langevin_dynamics}, which leads to relatively longer training time.

In the future, we would like to extend our approach to other saliency detection problems.
Also, we plan to capture new datasets with multiple human annotations to further model the statistics of human uncertainty in saliency perception.


 \vspace{-5pt}
 \section*{Acknowledgements.}
 \footnotesize{This research was supported in part by NSFC (61871325, 61671387, 61620106008, 61572264), National Key Research and Development Program of China (2018AAA0102803), 
and Tianjin Natural Science Foundation (17JCJQJC43700).}


%





\ifCLASSOPTIONcaptionsoff
  \newpage
\fi



%


\bibliographystyle{ieeetr}
\bibliography{saliency_ref}

\end{document}